%% file: main_stripped.tex
\documentclass[11pt]{article}
\usepackage[dvipsnames]{xcolor}

\usepackage{amsmath,amsbsy,amsfonts,amssymb,amsthm,fullpage}
\usepackage{graphicx}
\usepackage{tablefootnote}
\usepackage{url}
\usepackage{color,cases}
\usepackage{tikz}
\usetikzlibrary{calc,shapes}
\usepackage{subfigure}
\usepackage[margin=1in]{geometry}
\usepackage{xcolor}
\definecolor{DarkGreen}{rgb}{0.1,0.5,0.1}
\definecolor{DarkRed}{rgb}{0.5,0.1,0.1}
\definecolor{DarkBlue}{rgb}{0.1,0.1,0.5}
\usepackage{hyperref}
\hypersetup{
    unicode=false,              pdftoolbar=true,            pdfmenubar=true,            pdffitwindow=false,          pdfnewwindow=true,          colorlinks=true,           linkcolor=DarkBlue,              citecolor=DarkGreen,            filecolor=DarkGreen,          urlcolor=DarkBlue,                                      }
\usepackage{nicefrac}
\usepackage{algorithm}
\usepackage{algorithmic}
\usepackage{xspace}
\usepackage{wrapfig}

\input{macro}

\def\shownotes{0}  \ifnum\shownotes=1
\newcommand{\authnote}[2]{$\ll$\textsf{\footnotesize #1 notes: #2}$\gg$}
\else
\newcommand{\authnote}[2]{}
\fi
\newcommand{\Tnote}[1]{{\color{RedViolet}\authnote{Tengyu}{#1}}}

\newcommand{\FIXME}[1]{{\color{RedViolet}\authnote{Tengyu}{#1}}}

\title{On the Optimization Landscape of Tensor Decompositions}

\author{Rong Ge \thanks{Duke University, Computer Science Department, email: rongge@cs.duke.edu} \and Tengyu Ma\thanks{Princeton University, Computer Science Department, email: tengyu@cs.princeton.edu}}
\begin{document}
	\maketitle
\input{newintro_stripped}

\input{prelim_stripped}

\input{estimation_stripped}
\input{local_stripped}
\section{Conclusion} 
We analyze the optimization landscape of the random over-complete tensor decomposition problem using the Kac-Rice formula and random matrix theory.  We show that in the superlevel set $L$ that contains all the points with function values barely larger than the random guess, there are exactly $2n$ local maxima that correspond to the true components. This implies that with an initialization slight better than the random guess, local search algorithms converge to the desired solutions. We believe our techniques can be extended to 3rd order tensors, or other non-convex problems with structured randomness.

The immediate open question is whether there is any other spurious local maximum outside this superlevel set. Answering it seems to involve solving difficult questions in random matrix theory. Another potential approach to unravel the mystery behind the success of the non-convex methods is to analyze the early stage of local search algorithms and show that they will enter the superlevel set $L$ quickly.

\section*{Acknowledgements} 

This paper was done in part while the authors were hosted by Simons Institute. We are indebted to Ramon van Handel for very useful comments on random matrix theory. We thank Nicolas Boumal and Daniel Stern for helpful discussions. This research was supported by NSF, Office of Naval Research, and the Simons Foundation.

\appendix
\bibliographystyle{alpha}
\bibliography{full,ref_tensor_2}

\input{manifold_opt_stripped}

\input{toolbox_stripped}
\input{rip_stripped}

\input{proofproofoverview_stripped}

\end{document}

%% file: macro.tex
\newtheorem{theorem}{Theorem}[section]

\newtheorem{lemma}[theorem]{Lemma}
\newtheorem{claim}[theorem]{Claim}
\newtheorem{corollary}[theorem]{Corollary}

\newtheorem{proposition}[theorem]{Proposition}
\theoremstyle{definition}

\newtheorem{definition}[theorem]{Definition}

\numberwithin{equation}{section}

\newcommand{\IGNORE}[1]{}

\newcommand{\citep}{\cite}
\newcommand{\citet}{\cite}

\newcommand\E{\mathbb{E}}

\DeclareMathOperator{\diag}{diag}

\newcommand\inner[1]{\langle #1 \rangle}

\newcommand{\Exp}{\mathop{\mathbb E}\displaylimits}

\newcommand{\fourmoments}[1]{\Norm{#1}_4^4}
\newcommand{\sixmoments}[1]{\Norm{#1}_6^6}

\newcommand{\sphere}{S}

\newcommand{\cB}{\mathcal{B}}
\newcommand{\cM}{\mathcal{M}}
\newcommand{\tangent}{T}
\newcommand{\Diff}{\nabla}
\newcommand{\indicator}[1]{\mathbf{1}(#1)}
\newcommand{\Vol}{\textup{Vol}}
\newcommand{\bartau}{\bar{\tau}}
\newcommand{\allones}{\mathbf{1}}

\newcommand{\argmax}{\textup{argmax}}

\newcommand{\mathand}{~~\textup{and}~~}

\newcommand{\Set}[1]{{\left\{#1\right\}}}
\newcommand{\hessian}{\textup{Hess }}
\newcommand{\grad}{\textup{grad }}
\newcommand{\norm}[1]{\lVert#1\rVert}
\newcommand{\Norm}[1]{\left\lVert#1\right\rVert}
\newcommand{\Id}{\textup{Id}}

\newcommand{\mper}{\,.}
\newcommand{\mcom}{\,,}

\newcommand\R{\mathbb{R}}

\newcommand\N{\mathcal{N}}

\newcommand{\trace}{\textup{tr}}

\let\epsilon=\varepsilon

\numberwithin{equation}{section}

\newcommand\MYcurrentlabel{xxx}

\newcommand{\MYstore}[2]{%
	\global\expandafter \def \csname MYMEMORY #1 \endcsname{#2}%
}

\newcommand{\MYload}[1]{%
	\csname MYMEMORY #1 \endcsname%
}

\newcommand{\MYnewlabel}[1]{%
	\renewcommand\MYcurrentlabel{#1}%
	\MYoldlabel{#1}%
}

\newcommand{\MYdummylabel}[1]{}

\newcommand{\torestate}[1]{%
	\let\MYoldlabel\label%
	\let\label\MYnewlabel%
	#1%
	\MYstore{\MYcurrentlabel}{#1}%
	\let\label\MYoldlabel%
}

\newcommand{\restatetheorem}[1]{%
	\let\MYoldlabel\label
	\let\label\MYdummylabel
	\begin{theorem*}[Restatement of \prettyref{#1}]
		\MYload{#1}
	\end{theorem*}
	\let\label\MYoldlabel
}

\newcommand{\restatelemma}[1]{%
	\let\MYoldlabel\label
	\let\label\MYdummylabel
	\begin{lemma*}[Restatement of \prettyref{#1}]
		\MYload{#1}
	\end{lemma*}
	\let\label\MYoldlabel
}

\newcommand{\restateprop}[1]{%
	\let\MYoldlabel\label
	\let\label\MYdummylabel
	\begin{proposition*}[Restatement of \prettyref{#1}]
		\MYload{#1}
	\end{proposition*}
	\let\label\MYoldlabel
}

\newcommand{\restatefact}[1]{%
	\let\MYoldlabel\label
	\let\label\MYdummylabel
	\begin{fact*}[Restatement of \prettyref{#1}]
		\MYload{#1}
	\end{fact*}
	\let\label\MYoldlabel
}

\newcommand{\restate}[1]{%
	\let\MYoldlabel\label
	\let\label\MYdummylabel
	\MYload{#1}
	\let\label\MYoldlabel
}

%% file: newintro_stripped.tex
\pagenumbering{gobble}
\begin{abstract}
	\normalsize
	Non-convex optimization with local search heuristics has been widely 	used in machine learning, achieving many state-of-art results. 	It becomes increasingly important to understand why they can work for these NP-hard problems on typical data. The landscape of many objective functions in learning has been conjectured to have the geometric property that ``all local optima are (approximately) global optima'', and thus they can be solved efficiently by local search algorithms. 	However, establishing such property can be very difficult. 
	
	In this paper, we analyze the optimization landscape of the random over-complete  tensor decomposition problem, which has many applications in unsupervised learning, especially in learning latent variable models. In practice, it can be efficiently solved by gradient ascent on a non-convex objective. 
	We show that for any small constant $\epsilon > 0$, among the set of points with function values $(1+\epsilon)$-factor larger than the expectation of the function, all the local maxima are approximate global maxima. Previously, the best-known result only characterizes the geometry in small neighborhoods around the true components. 
	Our result implies that even with an initialization that is barely better than the random guess, the gradient ascent algorithm is guaranteed to solve this problem. 
	
	Our main technique uses Kac-Rice formula and random matrix theory. To our best knowledge, this is the first time when Kac-Rice formula is successfully applied to counting the number of local minima of a highly-structured random polynomial with dependent coefficients. \end{abstract}

\newpage
\pagenumbering{arabic}
\section{Introduction}

Non-convex optimization is the dominating algorithmic technique behind many state-of-art results in machine learning, computer vision, natural language processing and reinforcement learning. Local search algorithms through stochastic gradient methods are simple, scalable and easy to implement. Surprisingly, they also return high-quality solutions for practical problems like training deep neural networks, which are NP-hard in the worst case. 
It has been conjectured~\cite{dauphin2014identifying,choromanska2015loss} that on \textit{typical} data, the landscape of the training objectives has the nice geometric property that \textit{all local minima are (approximate) global minima}.  Such property assures the local search algorithms to converge to global minima~\cite{ge2015escaping,DBLP:conf/colt/LeeSJR16,nesterov2006cubic,sun2015nonconvex}. However, establishing it for concrete problems can be challenging. 

Despite recent progress on understanding the optimization landscape of various machine learning problems (see~\cite{ge2015escaping, bandeira2016low,  bhojanapalli2016global, Kawaguchi, ge2016matrix,DBLP:journals/corr/HardtM16,DBLP:journals/corr/HardtMR16} and references therein), a comprehensive answer remains elusive. 
Moreover, all previous techniques fundamentally rely on the spectral structure of the problems. For example,  in~\cite{ge2016matrix} 
allows us to pin down the set of the critical points (points with vanishing gradients) as approximate eigenvectors of some matrix. Among these eigenvectors we can further identify all the local minima. 
The heavy dependency on linear algebraic structure limits the generalization to problems with non-linearity (like neural networks).

Towards developing techniques beyond linear algebra, in this work, we investigate the optimization landscape of tensor decomposition problems. This is a clean non-convex optimization problem whose optimization landscape cannot be analyzed by the previous approach. It also connects to the training of neural networks with many shared properties~\cite{2015arXiv150906569N} . For example, in comparison with the matrix case where all the global optima reside on a (connected) Grassmannian manifold, for both tensors and neural networks all the global optima are isolated from each other. 

Besides the technical motivations above, tensor decomposition itself is also the key algorithmic tool for learning many latent variable models, mixture of Gaussians, hidden Markov models, dictionary learning~\cite{Chang96,MR06,HKZ12,AHK12,SpectralLDA,HK13-mog}, just to name a few. In practice,  local search heuristics such as alternating least squares~\cite{comon2009tensor}, gradient descent and power method~\cite{kolda2011shifted} are popular and successful. 

Concretely, we consider decomposing a random 4-th order tensor $T$ of the rank $n$ of the following form, 
$$
T = \sum_{i=1}^n a_i \otimes a_i \otimes a_i \otimes a_i\mper
$$

We are mainly interested in the \textit{over-complete} regime where $n \gg d$. This setting is particularly challenging, but it is crucial for unsupervised learning applications where the hidden representations have higher dimension than the data~\cite{rgDict2, de2007fourth}.  Previous algorithmic results either require access to  high order tensors \cite{bhaskara2014smoothed, fourierpca}, or use complicated techniques such as FOOBI~\cite{de2007fourth} or sum-of-squares relaxation~\cite{DBLP:conf/stoc/BarakKS15,ge2015decomposing, DBLP:conf/stoc/HopkinsSSS16,MSS16}.

In the worst case, most tensor problems are NP-hard~\cite{haastad1990tensor,hillar2013most}. Therefore we work in the average case where vectors $a_i \in \R^d$ are assumed to be drawn i.i.d from Gaussian distribution $\N(0, I)$. We call $a_i$'s the components of the tensor. We are given the entries of tensor $T$ and our goal is to recover the components $a_1,\dots, a_n$.  

We will analyze the following popular non-convex objective, 
\begin{align}
\max & \quad f(x) = \sum_{i,j,k,l\in [d]^4} T_{i,j,k,l} x_ix_jx_kx_l = \sum_{i=1}^n\inner{a_i,x}^4\label{eq:obj}\\
s.t. & \quad \|x\| = 1.\nonumber
\end{align}

It is known that for $n \ll d^2$, the global maxima of $f$ is close to one of $\pm \frac{1}{\sqrt{d}}a_1,\dots,\pm \frac{1}{\sqrt{d}}a_n$. Previously, Ge et al.~\cite{ge2015escaping} show that for the orthogonal case where $n \le d$ and all the $a_i$'s are \textit{orthogonal}, objective function $f(\cdot)$ have only $2n$ local maxima that are approximately $\pm \frac{1}{\sqrt{d}}a_1,\dots,\pm \frac{1}{\sqrt{d}}a_n$. However, the technique heavily uses the orthogonality of the components and is not generalizable to over-complete case. 

Empirically, projected gradient ascent and power methods find one of the components $a_i$'s even if $n$ is significantly larger than $d$.
The local geometry for the over-complete case around the true components is known: in a small neighborhood of each of  $\pm \frac{1}{\sqrt{d}}a_i$'s,  there is a unique local maximum~\cite{anandkumar2015learning}. Algebraic geometry techniques~\cite{cartwright2013number,2015arXiv150505729A} can show that $f(\cdot)$ has an exponential number of other critical points, while these techniques seem difficult to extend to the characterization of local maxima.
It remains a major open question whether there are any other spurious local maxima that gradient ascent can potentially converge to.

\paragraph{Main results.}

We show that there are no spurious local maxima in a large superlevel set that contains all the points with function values slightly larger than that of the random initialization. 

\begin{theorem}\label{thm:main_intro} Let $\epsilon, \zeta \in (0,1/3)$ be two arbitrary constants and $d$ be sufficiently large.   Suppose $d^{1+\epsilon} < n < d^{2-\epsilon}$. Then, with high probability over the randomness of $a_i$'s, we have that  in the superlevel set 
	\begin{align}
	L = \left\{x\in \sphere^{d-1}: f(x) \ge 3(1+\zeta)n\right\} \mcom\label{eqn:def-L-intro}
	\end{align}
	there are exactly $2n$ local maxima with function values $(1\pm o(1))d^2$, each of which is $\widetilde{O}(\sqrt{n/d^3})$-close to one of $\pm \frac{1}{\sqrt{d}}a_1,\dots, \pm \frac{1}{\sqrt{d}}a_n$. 
\end{theorem}
Previously, the best known result~\cite{anandkumar2015learning} only characterizes the geometry in small neighborhoods around the true components, that is, there exists one local minima in each of the small constant neighborhoods around each of the true components $a_i$'s. (It turns out in such neighborhoods, the objective function is actually convex.) We significantly enlarge this region to the superlevel set $L$, on which the function $f$ is not convex and has an exponential number of saddle points, but still doesn't have any spurious local minima. 

Note that a random initialization $z$ on the unit sphere has expected function value $\E[f(z)] = 3n$. Therefore the superlevel set $L$ contains all points that have function values barely larger than that of the random guess. 
Hence, Theorem~\ref{thm:main_intro} implies that with a slightly better initialization than the random guess, gradient ascent and power method\footnote{Power method is exactly equivalent to gradient ascent with a properly chosen finite learning rate} are guaranteed to find one of the components in polynomial time. (It is known that after finding one component, it can be peeled off from the tensor and the same algorithm can be repeated to find all other components.)

\begin{corollary}\label{cor:main_intro}
	In the setting of Theorem~\ref{thm:main_intro}, with high probability over the choice of $a_i$'s, we have that given any starting point $x^0$ that satisfies $f(x^0) \ge 3(1+\zeta)n$, stochastic projected gradient descent\footnote{We note that by stochastic gradient descent we meant the algorithm that is analyzed in~\cite{ge2015escaping}. To get a global minimizer in polynomial time (polynomial in $\log(1/\epsilon)$ to get $\epsilon$ precision), one also needs to slightly modify stochastic gradient descent in the following way: one can run SGD until $1/d$ accuracy and then switch to gradient descent. Since the problem is locally strongly convex, the local convergence is linear. } will find one of the $\pm \frac{1}{\sqrt{d}}a_i$'s up to $\widetilde{O}(\sqrt{n/d^3})$ Euclidean error in polynomial time. 
\end{corollary}

We also strengthen Theorem~\ref{thm:main_intro} and Corollary~\ref{cor:main_intro} (see Theorem~\ref{thm:main}) slightly -- the same conclusion still holds with  $\zeta = O(\sqrt{d/n})$ that is smaller than a constant. Note that the expected value of a random initialization is $3n$ and we only require an initialization that is slightly better than random guess in function value. We also \textbf{conjecture} that from random initialization, it suffices to use constant number of projected gradient descent (with optimal step size) to achieve the function value $3(1+\zeta)n$ with $\zeta = O(\sqrt{d/n})$. This conjecture --- an interesting question for future work --- is based on the hypothesis that the first constant number of steps of gradient descent can make similar improvements as the first step does (which is equal to $c\sqrt{dn}$ for a universal constant $c$).

As a comparison, previous works such as \cite{anandkumar2015learning} require an initialization with function value $\Theta(d^2) \gg n$. Anandkumar et al.~\cite{anandkumar2016analyzing} analyze the dynamics of tensor power method with a delicate initialization that is \textit{independent} with the randomness of the tensor. Thus it is not suitable for the situation where the initialization comes from the result of another algorithm, and it does not have a direct implication on the landscape of $f(\cdot)$.

We note that the local maximum of $f(\cdot)$ corresponds to the robust eigenvector of the tensor. Using this language, our theorem says that a robust eigenvector of an over-complete  tensor with random components is either one of those true components or has a small correlation with the tensor in the sense that $\inner{T,x^{\otimes 4}}$ is small. This improves significantly upon the understanding of robust eigenvectors~\cite{2015arXiv150505729A} under an interesting random model. 

\paragraph{Our techniques}

The proof of Theorem~\ref{thm:main_intro} uses Kac-Rice formula (see, e.g.,~\cite{adler2009random}), which is based on a counting argument. To build up the intuition,  we tentatively view the unit sphere as a collection of discrete points, then for each point $x$ one can compute the probability (with respect to the randomness of the function) that $x$ is a local maximum. Adding up all these probabilities will give us the expected number of local maxima. In continuous space, such counting argument has to be more delicate since the local geometry needs to be taken into account. This is formalized by Kac-Rice formula (see Lemma~\ref{lem:kac-rice}). 

However, Kac-Rice formula only gives a closed form expression that involves the integration of the expectation of some complicated random variable. It's often very challenging to simplify the expression to obtain interpretable results. 
Before our work, Auffinger et al.~\cite{auffinger2013random, auffinger2013complexity} have successfully applied Kac-Rice formula to characterize the landscape of polynomials with random Gaussian coefficients. The exact expectation of the number of local minima can be computed there, because the Hessian of a random polynomial is a Gaussian orthogonal ensemble, whose eigenvalue distribution is well-understood with closed form expression.

Our technical contribution here is successfully applying Kac-Rice formula to \textit{structured} random non-convex functions where the formula cannot be exactly evaluated.   The Hessian and gradients of $f(\cdot)$ have much more complicated distributions compared to the Gaussian orthogonal ensemble. As a result, the Kac-Rice formula is impossible to be evaluated exactly. We instead cut the space $\R^d$ into regions and use different techniques to \textit{estimate} the number of local maxima.  
See a proof overview in Section~\ref{sec:proof_sketches}. We believe our techniques can be extended to 3rd order tensors and can shed light on the analysis of other non-convex problems with structured randomness.

\paragraph{Organization} In Section~\ref{sec:prelim} we introduce preliminaries regarding manifold optimization and Kac-Rice formula. 
We give a detailed explanation of our proof strategy in Section~\ref{sec:proof_sketches}. We fill in the technical details in the later sections: in Section~\ref{sec:global} we show that there is no local maximum that is uncorrelated with any of the true components. We compliment that by a local analysis in Section~\ref{sec:local} that shows there are exactly $2n$ local optima around the true components.

%% file: prelim_stripped.tex
\section{Notations and Preliminaries}\label{sec:prelim}

We use $\Id_d$ to denote the identity matrix of dimension $d\times d$, and for a subspace $K$, let $\Id_K$ denote the projection matrix to the subspace $K$. For unit vector $x$, let $P_x = \Id - xx^{\top}$ denote the projection matrix to the subspace \textit{orthogonal} to $x$. 
Let $\sphere^{d-1}$ be the $d-1$-dimensional sphere $\sphere^{d-1} := \{x\in \R^d: \|x\|^2 = 1  \}$. 

Let $u \odot v$ denote the Hadamard product between vectors $u$ and $v$. Let $u^{\odot s}$ denote $u\odot \dots \odot u$ where $u$ appears $k$ times. Let $ A \otimes B$ denote the Kronecker product of $A$ and $B$. 
Let $\|\cdot\|$ denote the spectral norm of a matrix or the Euclidean norm of a vector. Let $\norm{\cdot}_F$ denote the Frobenius norm of a matrix or a tensor.

We write $A \lesssim B$ if there exists a universal constant $C$ such that $A\le CB$. We define $\gtrsim$ similarly. Unless explicitly stated otherwise, $O(\cdot)$-notation hides absolute multiplicative constants. Concretely, every occurrence of the notation $O(x)$ is a placeholder for some function $f(x)$ that satisfies $\forall x \in \R, |f (x)|\le   C|x|$ for some absolute constant $C > 0$. 
\subsection{Gradient, Hessian, and local maxima on manifold}

We have a constrained optimization problem over the unit sphere $\sphere^{d-1}$, which is a smooth manifold. Thus we define the local maxima with respect to the manifold. It's known that projected gradient descent for $\sphere^{d-1}$ behaves pretty much the same on the manifold as in the usual unconstrained setting~\cite{2016arXiv160508101B}. In Section~\ref{sec:manifold} we give a  brief introduction to manifold optimization, and the definition of gradient and Hessian.  We refer the readers to the book~\cite{absil2007optimization} for more backgrounds.
Here we use $\grad f$ and $\hessian f$ to denote the gradient and the Hessian of $f$ on the manifold $\sphere^{d-1}$. We compute them in the following claim. 

\begin{claim}\label{claim:grad-hessian}
	Let $f: S^{d-1}\rightarrow \R$ be $
	f(x) := \frac{1}{4}\sum_{i=1}^{n} \inner{a_i,x}^4
	$. 
	Then the gradient and Hessian of $f$ on the sphere can be written as, 
	\begin{align}
	\grad f(x) &= P_x\sum_{i=1}^{n} \inner{a_i,x}^3  a_i \mcom\nonumber\\
	\hessian f(x) & = 3 \sum_{i=1}^{n}\inner{a_i,x}^2 P_xa_ia_i^{\top}P_x - \left(\sum_{i=1}^{n} \inner{a_i,x}^4\right) P_x \mcom\nonumber
	\end{align}
	where $P_x = \Id_d - xx^{\top}$. 
\end{claim}
A local maximum of a function $f$ on the manifold $\sphere^{d-1}$ satisfies $\grad f(x) = 0$, and $\hessian f(x) \preceq 0$. Let $\cM_f$ be the set of all local maxima, 
\begin{align}
\cM_f = \Set{x\in \sphere^{d-1}: \grad f(x) = 0, \hessian f(x) \preceq 0} \mper \label{eqn:def-L}
\end{align}
\subsection{Kac-Rice formula}

Kac-Rice formula is a general tool for computing the expected number of special points on a manifold. Suppose there are two random functions $P(\cdot ): \R^d\rightarrow \R^d$ and $Q(\cdot): \R^d \rightarrow \R^k$, and an open set $\mathcal{B}$ in $\R^k$. The formula counts the expected number of point $x\in \R^d$ that satisfies both $P(x) =0 $ and $Q(x)\in \mathcal{B}$. 

Suppose we take $P = \nabla f$ and $Q = \nabla^2 f$, and let $\cB$ be the set of negative semidefinite matrices, then the set of points that satisfies $P(x) = 0$ and $Q \in \cB$ is the set of all local maxima $\cM_f$.  Moreover, for any set $Z\subset\sphere^{d-1}$, we can also augment $Q$ by $Q = [\nabla^2 f, x]$ and choose $\cB = \{A: A\preceq 0\}\otimes Z$. With this choice of $P,Q$, Kac-Rice formula can count the number of local maxima inside the region $Z$. For simplicity, we will only introduce Kac-Rice formula for this setting. We refer the readers to~\cite[Chapter 11\&12]{adler2009random} for more backgrounds. 
\begin{lemma}[Informally stated]\label{lem:kac-rice}
	Let $f$ be a random function defined on the unit sphere $\sphere^{d-1}$ and let $Z\subset \sphere^{d-1}$. Under certain regularity conditions\footnote{We omit the long list of regularity conditions here for simplicity. See more details at ~\cite[Theorem 12.1.1]{adler2009random}} on $f$ and $Z$, we have  
	\begin{align}
	\Exp\left[\cM_f\cap Z|\right] = \int_x \Exp\left[|\det(\hessian f)|\cdot \indicator{\hessian f\preceq 0} \indicator{x\in Z}\mid \grad f(x) = 0\right] p_{\grad f(x)}(0) dx\mper \label{eqn:16}
	\end{align}
	where $dx$ is the usual surface measure on $S^{d-1}$ and $p_{\grad f(x)}(0)$ is the density of $\grad f(x)$ at 0. 
\end{lemma}

\subsection{Formula for the number of local maxima}

In this subsection, we give a concrete formula for the number of local maxima of our objective function~\eqref{eq:obj} inside the superlevel set $L$ (defined in equation~\eqref{eqn:def-L-intro}). Taking $Z = L$ in Lemma~\ref{lem:kac-rice}, it boils down to estimating the quantity on the right hand side of~\eqref{eqn:16}. We remark that for the particular function $f$ as defined in~\eqref{eq:obj} and $Z = L$, the integrand in~\eqref{eqn:16} doesn't depend on the choice of $x$. This is because for any $x\in \sphere^{d-1}$, $(\hessian f, \grad f,\indicator{x\in L})$ has the same joint distribution, as characterized below: 

\begin{lemma}\label{lem:dist}
	Let $f$ be the random function defined in~\eqref{eq:obj}. Let $\alpha_1,\dots, \alpha_n \in \N(0,1)$, and $b_1,\dots, b_n \sim \N(0,\Id_{d-1})$ be independent Gaussian random variables. Let \begin{align}
	M &= \norm{\alpha}_4^4\cdot \Id_{d-1} -3\sum_{i=1}^{n} \alpha_i^2 b_ib_i^{\top} \mathand g  = \sum_{i=1}^{n}\alpha_i^3 b_i\label{eqn:def:g}
	\end{align} 
	Then, we have that for any $x\in \sphere^{d-1}$, $(\hessian f, \grad f, f)$ has the same joint distribution as $(-M,g, \fourmoments{\alpha})$.

\end{lemma}

\begin{proof}
	We use Claim~\ref{claim:grad-hessian}. 	We fix $x\in \sphere^{d-1}$ and let $\alpha_i = \inner{a_i,x}$ and $b_i = P_x a_i$. We have $\alpha_i$ and $b_i$ are independent, and $b_i$ is spherical Gaussian random vector in the tangent space at $x$ (which is isomorphic to $\R^{d-1}$). We can verify that $\grad f(x) = \sum_{i\in [n]} \alpha_i^3b_i = g$ and $\hessian f(x) =  -M$, and this complete the proof. 
\end{proof}

Using Lemma~\ref{lem:kac-rice} (with $Z=L$) and Lemma~\ref{lem:dist}, we derive the following formula for the expectation of our random variable $\Exp\left[|\cM_f\cap L|\right]$. Later we will later use Lemma~\ref{lem:kac-rice} slightly differently with another choice of $Z$. 
\begin{lemma}\label{lem:kac-rice-our-fun}
	Using the notation of Lemma~\ref{lem:dist}, let $p_g(\cdot)$ denote the density of $g$. Then, 
	\begin{align}
	\Exp\left[|\cM_f\cap L|\right] = \Vol(\sphere^{d-1})\cdot \Exp\left[\left|\det(M)\right|\indicator{M\succeq 0}\indicator{\fourmoments{\alpha}\ge 3(1+\zeta)n}\mid g=0\right] p_g(0) \mper \label{eqn:10}
	\end{align}
\end{lemma}

\section{Proof Overview}\label{sec:proof_sketches}

\newcommand{\Ezero}{\indicator{E_0}}
\newcommand{\Eone}{\indicator{E_1}}
\newcommand{\Etwo}{\indicator{E_2}}
\newcommand{\Fzero}{\indicator{F_0}}
\newcommand{\Fk}{\indicator{F_k}}

In this section, we give a high-level overview of the proof of the main Theorem. We will prove a slightly stronger version of Theorem~\ref{thm:main_intro}. 

Let $\gamma$ be a universal constant that is to be determined later. Define the set $L_1\subset \sphere^{d-1}$ as, 
\begin{align}
L_1 := \left\{x\in \sphere^{d-1}: \sum_{i=1}^n\inner{a_i,x}^4  \ge 3n + \gamma \sqrt{nd}\right\}\mper\label{eqn:def-L1}
\end{align}
Indeed we see that $L$ (defined in~\eqref{eqn:def-L-intro}) is a subset of $L_1$ when $n \gg d$. We prove that in $L_1$ there are exactly $2n$ local maxima. 
\begin{theorem}[main]\label{thm:main}
	There exists universal constants $\gamma,\beta$ such that the following holds: suppose $d^{2}/\log^{O(1)}\ge n \ge \beta d\log^2 d$ and $L_1$ be defined as in~\eqref{eqn:def-L1}, then with high probability over the choice of $a_1,\dots, a_n$, we have that the number of local maxima in $L_1$ is exactly $2n$: 
	\begin{align}
	|\cM_f\cap L_1| = 2n\mper\label{eqn:12}
	\end{align}
	Moreover, each of the local maximum in $L_1$ is $\widetilde{O}(\sqrt{n/d^3})$-close to one of $\pm \frac{1}{\sqrt{d}}a_1,\dots ,\pm\frac{1}{\sqrt{d}}a_n$. 
	\end{theorem}

\sloppy
In order to count the number of local maxima in $L_1$, we use the Kac-Rice formula (Lemma~\ref{lem:kac-rice-our-fun}). Recall that what Kac-Rice formula  gives an expression that involves the complicated expectation $\Exp\left[\left|\det(M)\right|\indicator{M\succeq 0}\indicator{\fourmoments{\alpha}\ge 3(1+\zeta)n}\mid g=0\right] $.  Here the difficulty is to deal with the determinant of a random matrix $M$ (defined in Lemma~\ref{lem:dist}), whose eigenvalue distribution does not admit an analytical form. Moreover, due to the existence of the conditioning and the indicator functions, it's almost impossible to compute the RHS of the Kac-Rice formula (equation~\eqref{eqn:10}) exactly.

\newcommand{\cC}{\mathcal{C}}

\paragraph{Local vs. global analysis} The key idea to proceed is to divide the superlevel set $L_1$ into two subsets 
\begin{align}
L_1 & =  (L_1\cap L_2) \cup L_2^c , \nonumber\\ \label{eq:L1L2}
& \textup{ where } L_2 := \{x\in \sphere^{d-1}: \forall i, \norm{P_x a_i}^2 \ge (1-\delta)d, \mathand |\inner{a_i,x}|^2\le \delta d\}\mper \end{align}
Here $\delta$ is a sufficiently small universal constant that is to be chosen later. We also note that $L_2^c\subset L_1$ and hence $L_1 =(L_1\cap L_2) \cup L_2^c $. 

Intuitively, the set $L_1\cap L_2$ contains those points that do not have large correlation with any of the $a_i$'s; the compliment $L_2^c$ is the union of the neighborhoods around each of the desired vector $\frac{1}{\sqrt{d}}a_1,\dots, \frac{1}{\sqrt{d}}a_n$. We will refer to the first subset $L_1\cap L_2$ as the global region, and refer to the $L_2^c$ as the local region. 

We will compute the number of local maxima in sets $L_1\cap L_2$ and $L^c_2$ separately using different techniques.
We will show that with high probability $L_1\cap L_2$ contains zero local maxima using Kac-Rice formula (see Theorem~\ref{thm:kac-rice-zero}).  Then, we show that $L^c_2$ contains exactly $2n$ local maxima (see Theorem~\ref{thm:local}) using a different and more direct approach. 
\paragraph{Global analysis.}

The key benefit of have such division to local and global regions is that for the global region, we can avoid evaluating the value of the RHS of the Kac-Rice formula. Instead, we only need to have an \textit{estimate}: 
Note that the number of local optima in $L_1\cap L_2$,  namely $|\cM_f\cap L_1\cap L_2|$,  is an integer nonnegative random variable. Thus, if we can show its expectation $\Exp\left[|\cM_f\cap L_1\cap L_2|\right]$ is much smaller than $1$, then Markov's inequality implies that with high probability, the number of local maxima will be \textit{exactly} zero. Concretely, we will use Lemma~\ref{lem:kac-rice} with $Z = L_1\cap L_2$, and then estimate the resulting integral using various techniques in random matrix theory. It remains quite challenging even if we are only shooting for an estimate. see Theorem~\ref{thm:kac-rice-zero} for the exact statement and Section~\ref{subsec:kac-rice} for an overview of the analysis.

\paragraph{Local analysis.} In the local region $L_2^c$, that is, the neighborhoods of $a_1,\dots, a_n$, we will show there are {\em exactly} $2n$ local maxima. As argued above, it's almost impossible to get exact numbers out of the Kac-Rice formula since it's often hard to compute the complicated integral. Moreover, Kac-Rice formula only gives the expected number but not high probability bounds. However, here the observation is that the local maxima (and critical points) in the local region are well-structured. Thus, instead, we show that in these local regions, the gradient and Hessian of a point $x$ are dominated by the terms corresponding to components $\{a_i\}$'s that are highly correlated with $x$. The number of such terms cannot be very large (by restricted isometry property, see Section~\ref{sec:rip}). As a result, we can characterize the  possible local maxima explicitly, and eventually show there is exactly one local maximum in each of the local neighborhoods around $\{\pm \frac{1}{\sqrt{d}} a_i\}$'s.

\paragraph{Concentration properties of $a_i$'s.} Before stating the formal results regarding the local and global analysis, we have the following technical preparation. 
Since $a_i$'s are chosen at random, with small probability the tensor $T$ will behave very differently from the average instances. The optimization problem for such irregular tensors may have much more local maxima. To avoid these we will restrict our attention to the following event $G_0$, which occurs with high probability (over the randomness of $a_i$'s) :
\begin{align}
G_0 := 
&\Big\{ \forall x\in \sphere^{d-1}~ \textup{the followings hold: }\nonumber\\
& ~~~\forall U\subset [n] \textup{ with }|U| < \delta n/\log d, \sum_{i\in U} \inner{a_i,x}^2 \le (1+\delta)d \mcom\label{eqn:32}\\
& ~~~\sum_{i\in [n]}\inner{a_i,x}^6\ge 15(1-\delta)n\mcom\label{eqn:29}\\
& ~~~n - 3\sqrt{nd}\le \sum_{i\in [n]}\inner{a_i,x}^2 \le n + 3\sqrt{nd}\Big\}\mper\label{eqn:101}
\end{align}
Here $\delta$ is a small enough universal constant. Event $G_0$ summarizes common properties of the vectors $a_i$'s that we frequently use in the technical sections. Equation \eqref{eqn:32} is the restricted isometry property (see Section~\ref{sec:rip}) that ensures the $a_i$'s are not adversarially correlated with each other. Equation \eqref{eqn:29} lowerbounds the high order moments of $a_i$'s. Equation \eqref{eqn:101} bounds the singular values of the matrix $\sum_{i=1}^n a_ia_i^\top$. These conditions will be useful in the later technical sections.

Next, we formalize the intuition above as the two theorems below. Theorem~\ref{thm:kac-rice-zero} states there is no local maximum in $L_1\cap L_2$, whereas Theorem~\ref{thm:local} concludes that there are exactly $2n$ local maxima in $L_1\cap L_2^c$.

\begin{theorem}\label{thm:kac-rice-zero}
	There exists universal small constant $\delta \in (0,1)$ and universal constants $\gamma,\beta$ such that  for sets $L_1,L_2$ defined in equation~\eqref{eq:L1L2} and $ n \ge \beta d\log^2 d$, we have that the expected number of local maxima in $L_1\cap L_2$ is exponentially small:
	\begin{align}
	\Exp\left[ |\cM_f\cap L_1\cap L_2| \cdot \indicator{G_0}\right]\le 2^{-d/2}\mper\nonumber
	\end{align}
\end{theorem}

\begin{theorem}\label{thm:local}
	Suppose $1/\delta^2 \cdot d\log d \le n \le d^2/\log^{O(1)} d$. Then, with high probability over the choice $a_1,\dots, a_n$, we have, 
	\begin{align}
	|\cM_f\cap L_1\cap  L_2^c| = 2n\mper\label{eqn:thm:local}
	\end{align}
	Moreover, each of the point in $L\cap L_2^c$ is $\widetilde{O}(\sqrt{n/d^3})$-close to one of $\pm \frac{1}{\sqrt{d}}a_1,\dots ,\pm\frac{1}{\sqrt{d}}a_n$. 
\end{theorem}

\paragraph{Proof of the main theorem}

The following Lemma shows that event $G_0$ that we conditioned on is indeed a high probability event. The proof follows from simple concentration inequalities and is deferred to Section~\ref{sec:proof:proof_sketches}. 
\begin{lemma}\label{lem:G_0}Suppose $1/\delta^2 \cdot d\log^2 d \le n \le d^2/\log^{O(1)} d$. Then, 
		\begin{align}
	\Pr\left[G_0\right] \ge 1 - d^{-10} \nonumber\mper
	\end{align}
\end{lemma}
\noindent Combining Theorem~\ref{thm:kac-rice-zero}, Theorem~\ref{thm:local} and Lemma~\ref{lem:G_0}, we obtain Theorem~\ref{thm:main} straightforwardly. 
\begin{proof}[Proof of Theorem~\ref{thm:main}]
	\sloppy
	By Theorem~\ref{thm:kac-rice-zero} and Markov inequality, we obtain that $\Pr\left[|L\cap L_1\cap L_2|\indicator{G_0} < 1/2\right]\le 2^{-d/2+1}$. Since $|L\cap L_1\cap L_2|\indicator{G_0}$ is an integer value random variable, therefore, we get $\Pr\left[|L\cap L_1\cap L_2|\indicator{G_0} = 0\right]\le 2^{-d/2+1}$. Thus, using Lemma~\ref{lem:G_0}, Theorem~\ref{thm:local} and union bound,  with high probability, we have that $G_0$, $|L\cap L_1\cap L_2|\indicator{G_0} =0$ and equation~\eqref{eqn:thm:local} happen. Then, using Theorem~\ref{thm:local} and union bound, we concluded that $|L\cap L_1| = |L\cap L_1\cap  L_2^c| + |L\cap L_1\cap  L_2| = 2n$. 
	\end{proof}
In the next subsections we sketch the basic ideas behind the proof of of Theorem~\ref{thm:kac-rice-zero} and Theorem~\ref{thm:local}. Theorem~\ref{thm:kac-rice-zero} is the crux of this technical part of the paper. 
\subsection{Estimating the Kac-Rice formula for the global region}
\label{subsec:kac-rice}
The general plan to prove Theorem~\ref{thm:kac-rice-zero} is to use random matrix theory to estimate the RHS of the Kac-Rice formula.  We begin by applying Kac-Rice formula to our situation. 

We first note that by definition of $G_0$, 
\begin{align}
|\cM_f\cap L_1\cap L_2| \cdot \indicator{G_0}\le \left|\cM_f\cap L_1\cap L_2\cap L_G\right|\mcom\label{eqn:102}
\end{align}
where $L_G = \{x\in \sphere^{d-1}: x~\textup{satisfies equation ~\eqref{eqn:32},~\eqref{eqn:29} and~\eqref{eqn:101}} \}$. Indeed, when $G_0$ happens, then $L_G = \sphere^{d-1}$, and therefore equation~\eqref{eqn:102} holds. 

Thus it suffice to control $\Exp\left[\left|\cM_f\cap L_1\cap L_2\cap L_G\right|\right]$. We will use the Kac-Rice formula (Lemma~\ref{lem:kac-rice}) with the set $Z = \cM_f\cap L_1\cap L_2\cap L_G$.

\paragraph{Applying Kac-Rice formula. } The first step to apply Kac-Rice formula is to characterize the joint distribution of the gradient and the Hessian.  We use the notation of Lemma~\ref{lem:dist} for expressing the joint distribution of $(\hessian f, \grad f, \indicator{x\in Z})$.
For any fix $x\in \sphere^{d-1}$, let $\alpha_i = \inner{a_i,x}$ and $b_i = P_x a_i$ (where $P_x = \Id-xx^{\top}$) and $M = \norm{\alpha}_4^4\cdot \Id_{d-1} -3\sum_{i=1}^{n} \alpha_i^2 b_ib_i^{\top} \mathand g  = \sum_{i=1}^{n}\alpha_i^3 b_i$ as defined in~\eqref{eqn:def:g}. In order to apply Kac-Rice formula, we'd like to compute the joint distribution of the gradient and the Hessian. 
We have that 
\begin{align}
\textup{ $(\hessian f, \grad f, \indicator{x\in Z})$ has the same distribution as $(M, g, \indicator{E_0}\indicator{E_1}\indicator{E_2}\indicator{E_2'})$}\mcom \nonumber
\end{align}
where $E_0,E_1,E_2,E_2'$ are defined as follows (though these details here will only be important later). 
\begin{align}
E_0 = 
& \left\{\forall U\subset [n] \textup{ with }|U| < \delta n/\log d, \|\alpha_U\|^2 \le (1+\delta)d\mcom \label{eqn:RIP}\right.\\
& ~~~
\Norm{\alpha}_4^4 \ge 15(1-\delta)n\mcom\nonumber\\
& ~~~
~n-3\sqrt{nd}\le \|\alpha\|^2 \le n + 3\sqrt{nd}\Big\}\mper\nonumber
\end{align}
We see that the equations above correspond to equation~\eqref{eqn:32},~\eqref{eqn:29} and~\eqref{eqn:101} and event $\Ezero$ corresponds to the event $x\in L_G$. Similarly, the following event $E_1$ corresponds to $x\in L_1$. 
\begin{align}
E_1 = & \left\{ \norm{\alpha}_4^4 \ge 3n + \gamma\sqrt{nd}\right\} \mper\nonumber
\end{align}
Events $E_2$ and $E_2'$ correspond to the events that $x\in L_2$. We separate them out to reflect that $E_2$ and $E_2'$ depends the randomness of $\alpha_i$'s and $b_i$'s respectively. 
\begin{align}
E_2 &= \left\{\|\alpha\|_{\infty}^2\le \delta d\right\}\nonumber \\
E_2' &=  \left\{\forall i\in [n], \|b_i\|^2 \ge (1-\delta)d\right\}\mper\nonumber
\end{align}
\noindent Using Kac-Rice formula (Lemma~\ref{lem:kac-rice} with $Z=L_1\cap L_2\cap L_G$), we conclude that
\begin{align}
\Exp\left[\left|\cM_f\cap L_1\cap L_2\cap L_G\right|\right] =     \Vol(\sphere^{d-1})\cdot\Exp\left[\left|\det(M)\right|\indicator{M\succeq 0}\indicator{E_0}\indicator{E_1}\indicator{E_2}\indicator{E_2'}\mid g=0\right] p_g(0)\mper\label{eqn:110}
\end{align}
Next,  towards proving Theorem~\ref{thm:kac-rice-zero} we will estimate the RHS of the equation~\eqref{eqn:110} using various techniques. 
\paragraph{Conditioning on $\alpha$.} We observe that the distributions of the gradient $g$ and Hessian $M$ are fairly complicated. In particular, we need to deal with the interactions of $\alpha_i$'s (the components along $x$) and $b_i$'s (the components in the orthogonal subspace of $x$). Therefore, we use the law of total expectation to first condition on $\alpha$ and take expectation over the randomness of $b_i$'s, and then take expectation over $\alpha_i$'s. Here below the inner expectation of RHS of~\eqref{eqn:100} is with respect to the randomness of $b_i$'s and the outer one is with respect to $\alpha_i$'s.

\begin{lemma}\label{lem:conditioning}
	Using the notation of Lemma~\ref{lem:dist}, let $E$ denotes an event and 
	let $p_{g\mid \alpha}$ denotes the density of $g\mid \alpha$. Then, 
	\sloppy
	\begin{align}
	\Exp\left[\left|\det(M)\right|\indicator{M\succeq 0}\indicator{E}\mid g=0\right] p_g(0)  = \Exp\left[\Exp\left[\left|\det(M)\right|\indicator{M\succeq 0} \indicator{E}\mid g = 0,\alpha\right] p_{g\mid \alpha}(0)\right] \mper \label{eqn:100}
	\end{align}
\end{lemma}

\noindent For notional convenience we define $h(\cdot):\R^n\rightarrow \R$ as 
\begin{align}
h(\alpha) &:= \Vol(\sphere^{d-1})\Exp\left[\det(M)\indicator{M\succeq 0}\indicator{E_2'}\mid g = 0, \alpha\right] \indicator{E_0}\indicator{E_1}\indicator{E_2}  p_{g\mid \alpha}(0) \mper\nonumber
\end{align}
Using equation~\eqref{eqn:102}, the Kac-Rice formula (equation~\eqref{eqn:110}), and law of total expectation (Lemma~\ref{lem:conditioning}), we obtain straightforwardly the following Lemma which gives an explicit formula for the number of local maxima in $L_1\cap L_2$. We provide a rigorous proof in Section~\ref{sec:proof:proof_sketches} that verifies the regularity condition of Kac-Rice formula. 
\begin{lemma}\label{lem:using-kac-rice-zero}
	Let $h(\cdot)$ be defined as above. In the setting of this section, we have
	\begin{align}
	\Exp\left[|\cM_f\cap L_1\cap L_2| \cdot \indicator{G_0}\right]\le \Exp\left[\left|\cM_f\cap L_1\cap L_2\cap L_G\right|\right]= \Exp\left[h(\alpha)\right] \mper\nonumber
	\end{align}
\end{lemma}

\noindent We note that $p_{g\mid \alpha}(0)$ has an explicit expression since $g\mid \alpha$ is Gaussian. For the ease of exposition, we separate out the hard-to-estimate part from $h(\alpha)$, which we call $W(\alpha)$: 
\begin{align}
W(\alpha) := \Exp\left[\det(M)\indicator{M\succeq 0}\indicator{E_2'}\mid g = 0, \alpha\right]\indicator{E_0}\indicator{E_1}\indicator{E_2}   \label{eqn:def-Walpha}\mper
\end{align}

\noindent Therefore by definition, we have that 
\begin{align}
h(\alpha) = \Vol(\sphere^{d-1})  W(\alpha)  p_{g\mid \alpha}(0)\label{def:Wa}\mper\end{align}
\noindent Now, since we have conditioned on $\alpha$, the distributions of the Hessian, namely $M\mid \alpha$, is a generalized Wishart matrix which is slightly easier than before. However there are still several challenges that we need to address in order to estimate $W(\alpha)$. 

\paragraph{Question I: how to control $\det(M)\indicator{M\succeq 0}$?} Recall that $M = \fourmoments{\alpha} -3\sum \alpha_i^2b_ib_i^{\top}$, which is a generalized Wishart matrix whose eigenvalue distribution has no (known) analytical expression. The determinant itself by definition is a high-degree polynomial over the entries, and in our case, a complicated polynomial over the random variables $\alpha_i$'s and vectors $b_i$'s. We also need to properly exploit the presence of the indicator function $\indicator{M\succeq 0}$, 
since otherwise, the desired statement will not be true -- the function $f$ has an exponential number of critical points. 

Fortunately, in most of the cases, we can use the following simple claim that bounds the determinant from above by the trace. The inequality is close to being tight when all the eigenvalues of $M$ are similar to each other. More importantly, it uses naturally the indicator function $\indicator{M\succeq 0}$! Later we will see how to strengthen it when it's far from tight.

\begin{claim}\label{claim:amgm}
	For any $p\times p$ symmetric matrix $V$, we have, 
	\begin{align}
	\det(V)\indicator{V\succeq 0}& \le \left(\frac{|\trace(V)|}{p}\right)^p\indicator{V\succeq 0} \nonumber
	\end{align}
\end{claim}
The claim is a direct consequence of AM-GM inequality. 
\begin{proof}
	We can assume $V$ is positive semidefinite since otherwise both sides of the inequality vanish. Then, suppose $\lambda_1,\dots, \lambda_p \ge 0$ are the eigenvalues of $V$. We have $\det(V) = \lambda_1\dots \lambda_p$ and $|\trace(V)| = \trace(V) = \lambda_1+\dots +\lambda_p$. Then by AM-GM inequality we complete the proof. 
\end{proof}
Applying the Lemma above with $V = M$, we have
\begin{align}
W(\alpha)\le \Exp\left[\frac{|\trace(M)|^{d-1}}{(d-1)^{d-1}} \mid g = 0, \alpha\right]\indicator{E_0}\indicator{E_1} \mper \label{eqn:103}
\end{align}
Here we dropped the indicators for events $E_2$ and $E_2'$ since they are not important for the discussion below. It turns out that $|\trace(M)|$ is a random variable that concentrates very well, and thus we have $\Exp\left[|\trace(M)|^{d-1}\right] \approx |\Exp\left[\trace(M)\right]|^{d-1}$. It can be shown that (see Proposition~\ref{prop:det-to-trace} for the detailed calculation), 
\begin{align}
\Exp\left[\trace(M)\mid g=0,\alpha\right] = (d-1)\left(\|\alpha\|_4^4 - 3\|\alpha\|^2 + 3\norm{\alpha}_8^8/\|\alpha\|_6^6\right) \nonumber\mper
\end{align}
Therefore using equation~\eqref{eqn:103} and equation above, ignoring $\indicator{E_2'}$, we have that 
\begin{align}
W(\alpha)\le \left(\|\alpha\|_4^4 - 3\|\alpha\|^2 + 3\norm{\alpha}_8^8/\|\alpha\|_6^6\right)^{d-1}\indicator{E_0}\indicator{E_1} \mper\nonumber
\end{align}
Note that since $g\mid \alpha$ has Gaussian distribution, we have, 
\begin{align}
p_{g\mid \alpha}(0) = (2\pi)^{-d/2} (\Norm{\alpha}_6^6)^{-d/2} \mper\label{eqn:104}
\end{align}
Thus using two equations above, we can bound $\Exp\left[h(\alpha)\right] $ by 
\begin{align}
\Exp\left[h(\alpha)\right] \le \Vol(\sphere^{d-1}) \Exp\left[\left(\|\alpha\|_4^4 - 3\|\alpha\|^2 + 3\norm{\alpha}_8^8/\|\alpha\|_6^6\right)^{d-1}\cdot (2\pi)^{-d/2} (\Norm{\alpha}_6^6)^{-d/2} \indicator{E_0}\indicator{E_1}\right]\mper\label{eqn:105}
\end{align}
Therefore, it suffices to control the RHS of~\eqref{eqn:105}, which is much easier than the original Kac-Rice formula. However, we still need to be careful here, as it turns out that RHS of~\eqref{eqn:105} is roughly $c^d$ for some constant $c > 1$! Roughly speaking, this is because the high powers of a random variables is very sensitive to its tail. 
\paragraph{Easy case when all $\alpha_i$'s are small. } To find a tight bound for the RHS of \eqref{eqn:105}, intuitively we can consider two events: the event $F_0$ when all of the $\alpha_i$'s are close to constant (defined rigorously later in equation~\eqref{eqn:def-F0}) and the complementary event $F_0^c$. 

We claim that $\Exp\left[h(\alpha)\Fzero\right]$ can be bounded by $2^{-d/2}$. Then we will argue that it's difficult to get an upperbound for $\Exp\left[h(\alpha)\indicator{F_0^c}\right]$ that is smaller than 1 using the RHS of equation~\eqref{eqn:105}. 

It turns out in most of the calculations below, we can safely ignore the contribution of the term $3\norm{\alpha}_8^8/\|\alpha\|_6^6$. For notation convenience, let $Q(\cdot): \R^d \rightarrow \R$ be defined as: 
\begin{align}
Q(z) = \Norm{z}_4^4 - 3\Norm{z}^2 \label{eqn:def:Q}\mper
\end{align}
When conditioned on the event $F_0$, roughly speaking, random variable $Q(\alpha) = \|\alpha\|_4^4 - 3\|\alpha\|^2$ behaves like a Gaussian distribution with variance $\Theta(n)$, since $Q(\alpha)$ is a sum of  independent random variables with constant variances. Note that $\Eone$ and $\Ezero$ imply that $Q(\alpha)= \norm{\alpha}_4^4-3\norm{\alpha}_2^2\ge (\gamma-3) \sqrt{nd}$. Therefore, $Q(\alpha)\indicator{E_0}\indicator{E_1}$ behaves roughly like a truncated Gaussian distribution, that is, $X\cdot \indicator{X\ge (\gamma-3)\sqrt{nd}}$ where $X\sim \N(0,\Theta(\sqrt{n}))$.  Then for sufficiently large constant $\gamma$, 
\begin{align}
\Exp\left[Q(\alpha)^{d-1}\indicator{E_0}\Eone \Fzero\right]\le (0.1nd)^{d/2}\mper\label{eqn:106}
\end{align} 
Moreover, recall that when the event $E_0$ happens, we have that $\sixmoments{\alpha} \ge 15(1-\delta)n$. Therefore putting all these bounds together into equation~\eqref{eqn:105} with the indicator $\Fzero$,  and using the fact that $\Vol(\sphere^{d-1}) = \frac{\pi^{d/2}}{\Gamma(d/2+1)}$, we can obtain the target bound, 
\begin{align}
\Exp\left[h(\alpha)\Fzero\right] & \le \frac{\pi^{d/2}}{\Gamma(d/2+1)} \cdot (0.1nd)^{d/2} \cdot  (2\pi)^{-d/2} (15(1-\delta)n)^{-d/2}  \le 2^{-d/2} \mper\nonumber
\end{align}

\paragraph{The heavy tail problem: } Next we explain why it's difficult to achieve a good bound from using RHS~\eqref{eqn:105}. The critical issue is that the random variable $Q(\alpha)$ has very heavy tail, and exponentially small probability \textit{cannot} be ignored. To see this, we first claim that $Q(\alpha)$ has a tail distribution that roughly behaves like
\begin{align}
\Pr\left[Q(\alpha) \ge t\right] \ge \exp(-\sqrt{t}/2)\mper\nonumber
\end{align}
Taking $t=d^2$, then the contribution of the tail to the expectation of $Q(\alpha)^d$ is at least $\Pr\left[Q(\alpha) \ge t\right] \cdot t^d \approx d^{2d}$, which is much larger than what we obtained (equation~\eqref{eqn:106}) in the case when all $\alpha_i$'s are small.

The obvious fix is to consider $Q(\alpha)^d (\norm{\alpha}_6^6)^{-d/2}$ together (instead of separately). For example, we will use Cauchy-Schwarz inequality to obtain that $Q(\alpha)^d (\norm{\alpha}_6^6)^{-d/2} \le \norm{\alpha}^d$. However, this alone is still not enough to get an upper bound of RHS of~\eqref{eqn:105} that is smaller than 1. In the next paragraph, we will tighten the analysis by strengthening the estimates of $\det(M)\indicator{M\succeq 0}$. 

\Tnote{TODO: take a look at this again}
\paragraph{Question II: how to tighten the estimate of $\det(M)\indicator{M\succeq 0}$?}

It turns out that the AM-GM inequality is not always tight for controlling the quantity $\det(M)\indicator{M\succeq 0}$. In particular, it gives pessimistic bound when $M\succeq 0$ is not true. As a matter of fact, whenever $F_0^c$ happens,  $M$ is unlikely to be positive semidefinite! (The contribution of $ \det(M)\indicator{M\succeq 0}$ will not be negligible since there is still tiny chance that $M$ is PSD. As we argued before, even exponential probability event cannot be safely ignored.) We will show that formally the event $\indicator{M\succeq 0}\indicator{F_0^c}$ have small enough probability that can kill the contribution of $Q(\alpha)^d$ when it happens (see Lemma~\ref{lem:prob_indicator} formally). 

\paragraph{Summary with formal statements. }
Below, we formally setup the notation and state two Propositions that summarize the intuitions above. Let $\tau = Kn/d$ where $K$ is a universal constant that will be determined later. Let $F_k$ be the event that 
\begin{align}
&
F_0 = \Set{\norm{\alpha}_{\infty}^4 \le \tau} \label{eqn:def-F0}  \\
& F_k = \Set{k = \argmax_{i\in [n]} \alpha_i^4 \mathand \alpha_k^4 \ge \tau} \textup{ for } 1\le k \le n \label{eqn:def-Fk}
\end{align}
We note that $1\le \indicator{F_0} + \indicator{F_1}+ \dots + \indicator{F_k}$, and therefore we have 
\begin{align}
h(\alpha) \le \sum_{k=0}^n h(\alpha)\indicator{F_k} \label{eqn:71}
\end{align}
Therefore towards controlling $\Exp\left[h(\alpha)\right]$, it suffices to have good upper bounds on $\Exp\left[h(\alpha)\indicator{F_k}\right]$ for each $k$, which are summarized in the  following two propositions. 

\begin{proposition}\label{prop:f_k}
	Let $K\ge 2\cdot 10^3$ be a universal constant. 
	Let $\tau = Kn/d$ and let $\gamma, \beta$ be sufficiently large constants (depending on $K$). Then for any $n \ge \beta d\log^2 d$, we have  that for any $ k\in \{1,\dots, n\}$
	\begin{align}
	\Exp\left[h(\alpha)\indicator{F_k}\right] \le (0.3)^{d/2} \mper\nonumber
	\end{align}
\end{proposition}

\begin{proposition}\label{prop:f_0}
	In the setting of Proposition~\ref{prop:f_k}, we have, 
	\begin{align}
	\Exp\left[h(\alpha)\indicator{F_0}\right] \le (0.3)^{d/2}\mper\nonumber
	\end{align}
\end{proposition}
\noindent We see that Theorem~\ref{thm:kac-rice-zero} can be obtained as a direct consequence of Proposition~\ref{prop:f_0}, Proposition~\ref{prop:f_k} and Lemma~\ref{lem:using-kac-rice-zero}. 

\begin{proof}[Proof of Theorem~\ref{thm:kac-rice-zero}]
	Using equation~\eqref{eqn:71} and Lemma~\ref{lem:using-kac-rice-zero}, we have that 
	\begin{align}
	& \Exp\left[\left|L\cap L_1\cap L_2\right|\indicator{G_0}\right]  \le  \Exp\left[h(\alpha)\right] \le  \Exp\left[h(\alpha)\Fzero\right]+ \sum_{k=1}^n \Exp\left[h(\alpha)\Fk\right] \nonumber\\
	& \le (n+1)\cdot (0.3)^{d/2}\le 2^{-d/2}\mper\tag{by Propostion~\ref{prop:f_0} and~\ref{prop:f_k}}
	\end{align}
\end{proof}

\subsection{Local analysis}
For a point $x$ in the local region $L_2^c$, the gradient and the Hessian are dominated by the contributions from the components that have a large correlation with $x$. Therefore it is easier to analyze the first and second order optimality condition directly.

Note that although this region is small, there are still exponentially many critical points (as an example, there is a critical point near $\frac{a_1+a_2}{\|a_1+a_2\|}$). Therefore our proof still needs to consider the second order conditions and is more complicated than Anandkumar et al.~\cite{anandkumar2015learning}.

We first show that all local maxima in this region must have very high correlation with one of the components:

\begin{lemma}
	In the setting of Theorem~\ref{thm:local}, for any local maximum $x\in L_2$, there exists a component $a \in \{\pm a_i/\|a_i\|\}$ such that $\inner{x,a} \ge 0.99$.
\end{lemma}

Intuitively, if there is a unique component $a_i$ that is highly correlated with $x$, then the gradient $\grad f(x) = \sum_{j=1}^n \inner{a_j,x}^3 a_j$ will be even more correlated with $a_i$. On the other hand, if $x$ has similar correlation with two components (e.g., $x = \frac{a_1+a_2}{\|a_1+a_2\|}$), then $x$ is close to a saddle point, and the Hessian will have a positive direction that makes $x$ more correlated with one of the two components. 

Once $x$ has really high correlation with one component ($|\inner{x,a_i}| \ge 0.99\|a_i\|$), we can prove that the gradient of $x$ is also in the same local region. Therefore by fixed point theorem, there must be a point where $\grad f(x) = x$, and it is a critical point. We also show the function is strongly concave in the whole region where $|\inner{x,a_i}| \ge 0.99\|a_i\|$. Therefore we know the critical point is actually a local maximum.

%% file: estimation_stripped.tex
\section{Bounding the number of local minima using Kac-Rice Formula}\label{sec:global}

In this section we give the proof of Proposition~\ref{prop:f_0} and Proposition~\ref{prop:f_k}, using the intuition described in Section~\ref{sec:proof_sketches}. 
\subsection{Proof of Proposition~\ref{prop:f_0}}

Following the plan in Section~\ref{sec:proof_sketches}, we start with using AM-GM inequality for the determinant. 
\begin{lemma}\label{cor:det-to-trace-app}
	Under the setting of Proposition~\ref{prop:f_0}, we have, 
	\begin{align}
		W(\alpha)\indicator{F_0} \Ezero\Etwo\Etwo\le \Exp\left[\frac{|\trace(M)|^{d-1}}{(d-1)^{d-1}} \mid g = 0, \alpha\right] \indicator{F_0}\indicator{E_0} \Eone\Etwo \nonumber\mper
	\end{align}
\end{lemma}

The next thing that we need to bound is $\Exp\left[|\trace(M)|^{d-1} \mid g, \alpha\right]$. Here the basic intuition is that since $\trace(M)\mid g,\alpha$ is quite concentrated around its mean and therefore we can switch the $(d-1)$-th power with the expectation without losing much. Proving such a result would require the understanding of the distribution of $M\mid g,\alpha$. We have the following  generic Claim that computes this distribution in a more general setting. 

\begin{claim}\label{claim:remove_conditioning}
	Let $w,y\in \R^n$ be two fixed vectors, and let $\bar{q} = y^{\odot 3}/\norm{ y^{\odot 3}}$. 
	Let $B = [b_1,\dots, b_n]\in \R^{(d-1)\times n}$ be a matrix with independent standard Gaussians as entries. Let $M_w = \|w\|_4^4\cdot\Id_{d-1} -3\sum_{i=1}^{n}w_i^2 b_ib_i^{\top}$ and $g_y  = \sum_{i=1}^{n}y_i^3 b_i$. 
	Let $Z\in \R^{(d-1)\times n}$ be a random matrix with rows independent drawn from $\N(0,\Id_n - \bar{q}\bar{q}^{\top})$. 
	
	Then, we have that $B \mid (g_y = 0)$ has the same distribution as $Z$, and consequently $M_w\mid g_y = 0$ has the same distribution as 
	\begin{align}
\fourmoments{w}\Id_{d-1} - 3Z\diag(w^{\odot 2})Z^{\top}\mper\nonumber
	\end{align}
\end{claim}

One can see that the setting that we are mostly interested in corresponds to $w = y=\alpha$ in Claim~\ref{claim:remove_conditioning}. 
\begin{proof}
			Let $c_1^{\top},\dots, c_{d-1}^{\top}$ be the rows of $B$. We write $c_i =\beta_i\bar{q} + z_i$ where $\beta_i = \inner{c_i,\bar{q}}$ and $z_i = (\Id_n-\bar{q}\bar{q}^{\top})c_i$. Let $Z$ denote the matrix with $z_i^{\top}$ as rows. Therefore we have $B = Z + \beta \bar{q}^{\top}$. 	
	Recall that $b_i$ are independent Gaussians with covariance $\N(0,\Id_{d-1})$. Thus we have that $c_i$'s are independent Gaussians with covariance $\N(0,\Id_n)$, and consequently $\beta_i$ and $z_i$ are independent random variables with $\N(0, 1)$ and $\N(0,\Id_n - \bar{q}\bar{q}^{\top})$ respectively. Moreover, note that $g = 0$ is equivalent to that $\forall i\in [d-1], \inner{c_i,q} = 0$, which in turn is equivalent to that $\forall i\in [d-1], \beta_i = 0$. Hence, $B \mid (g_y = 0)$ has the same distribution as $Z$, and consequently $M_w\mid (g_y=0)$ has the same distribution as $\|w\|_4^4\Id_{d-1} - 3Z\diag(w^{\odot 2})Z^{\top}$. 
\end{proof}

Using Claim~\ref{claim:remove_conditioning}, we can compute the the expectation of the trace and its $p$-th moments. Again we state the following proposition in a more general setting for future re-usability. 
\begin{proposition}\label{prop:det-to-trace}
	In the setting of Claim~\ref{claim:remove_conditioning}, we have
	\begin{align}
	\Exp\left[\trace(M_w)\mid g_y\right] = (d-1)\left(\|w\|_4^4 - 3\|w\|^2 + 3\inner{y^{\odot 6},w^{\odot 2}}/\|y\|_6^6\right)\mper \nonumber
	\end{align}
	Moreover,  for any $\epsilon \in (0,1)$, 
	\begin{align}
	\Exp\left[|\trace(M_w)|^p\mid g_y = 0\right]\le \max\left\{(1+\epsilon)^p \left|\Exp\left[\trace(M_w)\mid g_y\right]\right|^p, (\frac{1}{\epsilon}+1)^p\max\{4p\|w\|_{\infty}^2, 2 \sqrt{pd}\|w\|_4^2\}^p\right\} \label{eqn:21}
	\end{align}
\end{proposition}

\begin{proof}[Proof of Proposition~\ref{prop:det-to-trace}]
We use the same notations as in the proof of Claim~\ref{claim:remove_conditioning}.	
By Claim~\ref{claim:remove_conditioning}, we have $B =[b_1,\dots, b_n] \mid (g_y = 0)$ has the same distribution as $Z$, and consequently $M_w\mid (g_y=0)$ has the same distribution as $\|w\|_4^4\Id_{d-1} - 3Z\diag(w^{\odot 2})Z^{\top}$. 
	Furthermore, this implies that $\trace(M_w \mid (g_y = 0))$ has the same distribution as the random variable (denoted by $\Gamma$ below)
	\begin{align}
	\Gamma := \trace\left(\|w\|^4\Id_{d-1} - 3Z\diag(w^{\odot 2})Z^{\top}\right) = (d-1)\|w\|_4^4 - 3\sum_{k\in [d-1]} \|z_k\odot w\|^2\mper
	\end{align}

Since $z_k\odot w$ is Gaussian random variable, we have that $\|z_k\odot w\|^2$ is a sub-exponential random variable with parameter $(\nu,\rho)$ satisfying  $\nu\lesssim \|w\|_4^2$ and $\rho\lesssim \|w\|_{\infty}^2$ (see Definition~\ref{def:sub-exponential} for the definition of sub-exponential random variable and see Lemma~\ref{lem:zwnorm} for the exact calculation of the parameters for this case) . It is well known that when one takes sum of independent sub-exponential random variable, the $\rho$ parameter will be preserved and $\nu^2$ (which should be thought of as variance) will added up (see Lemma~\ref{lem:sub-exp-addition} for details).   Therefore, we have that $\Gamma$ is a sub-exponential random variable with parameter $(\nu^*, \rho^*)$ satisfying $\nu^* \lesssim \sqrt{d-1}\|w\|_4^2$ and $\rho^*\lesssim \|w\|_{\infty}^2$. Moreover, by Lemma~\ref{lem:zwnorm} again and the linearity of expectation,  we can compute the expectation of $\Gamma$,  \begin{align}
	\Exp\left[\Gamma\right] = (d-1)\left(\|w\|_4^4 - 3\|w\|^2 + 3\inner{\bar{q}^{\odot 2},w^{\odot 2}}\right) \mper\label{eqn:8} 
	\end{align} 
Next the basic intuition is if $\nu^*$ and $\rho^*$ are small enough compared to the mean of $\Gamma$, then $\Gamma$ is extremely concentrated around its mean, and thus $\Exp\left[|\Gamma|^{p} \right]$ is close to $\Exp\left[\Gamma\right]^p$. Otherwise, the mean is more or less negligible compared to $\max\{p\norm{w}_{\infty}, 2\sqrt{pd}\norm{w}_4^2\}$,  then  the moment of $\Gamma$ behaves like the moments of an sub-exponential random variable with mean 0. 	Indeed, using basic integration, we can show (see Lemma~\ref{lem:sub-exp-before_truncation-moments}) that a sub-exponential random variable with mean $\mu$ and parameter $(\nu,b)$ satisfies that for any $\epsilon\in (0,1)$, 
\begin{align}
		\Exp |X|^p  \lesssim \frac{1}{\epsilon^{p-1}} |\mu|^p + \frac{1}{(1-\epsilon)^{p-1}} \left(\nu^p p!! + (2b)^p p!\right)\mper \nonumber
		\end{align}
		Applying the equation above to $\Gamma$ completes the proof. 
\end{proof}

Using Proposition~\ref{prop:det-to-trace} with $w = y=\alpha$. we obtain the following bounds for the moments of the trace of $M$. 

\begin{corollary}\label{cor:tracepower} For any fixed $\alpha$, we have
		\begin{align}
		\Exp\left[\left(\frac{|\trace(M)|}{d-1}\right)^{p}\mid g = 0, \alpha\right] \indicator{E_0}\indicator{E_2}
		& \lesssim (1+O(\delta))^{p} \max\left\{Q(\alpha)^p, (0.1d)^p, \left(0.1pd^{-1/2}n^{1/2}\right)^{p}\right\}\mper\nonumber
		\end{align}
\end{corollary}

\begin{proof}
	We fix the choice of $\alpha$ and assume it satisfies event $F_0$. We will apply Proposition~\ref{prop:det-to-trace} with $w = \alpha$ and $y = \alpha$. We can verify that when $E_0$ and $E_2$ both happen, we have
	\begin{align}
	\max\{4p\|w\|_{\infty}^2, 2\sqrt{pd}\norm{w}_4^2\} & \lesssim  \max\{p\|\alpha\|_{\infty}^2, p\norm{\alpha}_4^2\} \le p\norm{\alpha}_{\infty}\Norm{\alpha}_2\nonumber \\
	& \lesssim \delta pd^{1/2}n^{1/2}\mper\tag{since $\alpha$ satisfies event $E_0,E_2$}
	\end{align}
	Therefore, taking $\epsilon = O(\delta)$ in Proposition~\ref{prop:det-to-trace}, we have that 
	\begin{align}
	\Exp\left[|\trace(M)|^{p}\mid g = 0, \alpha\right] \indicator{E_0}\indicator{E_2}\lesssim (1+O(\delta))^{p} \max\left\{\mu^{p}, \left(0.1pd^{1/2}n^{1/2}\right)^{p}\right\} \nonumber\mcom
	\end{align}
	where $\mu = \left|\Exp\left[\trace(M)\right]\right| = (d-1) (\|\alpha\|^4 - 3\|\alpha\|^2 + \frac{3\|\alpha\|_8^8}{\|\alpha\|_6^6})$. When $E_0$ happens, we have $\mu \le (d-1)(\Norm{\alpha}_4^4-3\Norm{\alpha}^2 + 3\Norm{\alpha}_{\infty}^2) \le (d-1)(\Norm{\alpha}_4^4-3\Norm{\alpha}^2 + O(\delta d))$. It follows that 
	\begin{align}
	\Exp\left[\left(\frac{|\trace(M)|}{d-1}\right)^{p}\mid g = 0, \alpha\right] \indicator{E_0}\indicator{E_2}& \lesssim (1+O(\delta))^{p} \max\left\{(Q(\alpha) + O(\delta d))^p, \left(0.1pd^{-1/2}n^{1/2}\right)^{p}\right\}\mper\nonumber\\
& \lesssim (1+O(\delta))^{p} \max\left\{Q(\alpha)^p, (0.1d)^p, \left(0.1pd^{-1/2}n^{1/2}\right)^{p}\right\}\mper 
	\end{align}
\end{proof}

Putting Corollary~\ref{cor:tracepower} and Lemma~\ref{cor:det-to-trace-app} together, we can prove Proposition~\ref{prop:f_0}. 
\begin{proof}[Proof of Proposition~\ref{prop:f_0}]

	Moreover, when $E_0$ happens, we have, 
	\begin{align}
	p_{g\mid \alpha}(0) &= (2\pi)^{-d/2} (\Norm{\alpha}_6^6)^{-d/2} \label{eqn:41}\\
	& \le (1+O(\delta))^d (2\pi)^{-d/2} (15n)^{-d/2}\label{eqn:44}\mper
	\end{align}
	It follows that 
	\begin{align}
		\Exp\left[  W(\alpha)  p_{g\mid \alpha}(0)\right] &	\le (1+O(\delta))^d (30\pi n)^{-d/2} \Exp\left[\Exp\left[\left|\frac{\trace(M)}{d-1}\right|^{(d-1)}\mid g = 0, \alpha\right] \indicator{E_1}\indicator{E_2}\indicator{F_0}\right] \tag{by Corollary~\ref{cor:det-to-trace-app}}\\
	&\le (1+O(\delta))^d (30\pi n)^{-d/2}\Exp\left[\max\left\{Q(\alpha)^{d-1}, (0.1d)^{d-1}, \left(0.1d^{1/2}n^{1/2}\right)^{p}\right\}\indicator{E_1}\indicator{F_0}\right]\tag{by Corollary~\ref{cor:tracepower} with $p = d-1$}\\
	&\le (1+O(\delta))^d (30\pi n)^{-d/2}\Exp\left[\max\left\{Q(\alpha_S)^{d-1},  \left(0.1d^{1/2}n^{1/2}\right)^{p}\right\}\indicator{E_1}\right] \label{eqn:11}	\end{align}
			For simplicity, Let $Z_i = (\alpha_i^4-3\alpha_i^2)\indicator{|\alpha_i|\le \tau^{1/4}}$, and $Z = \sum_i Z_i$. We observe that $E_1$ and $E_0$ implies that $Z\ge (\gamma-3)\sqrt{nd}$. Let $\gamma'=\gamma-3$.
	Thus, $Z^d\indicator{Z\ge {\gamma'} \sqrt{nd}}\le (\sum_{i\in [n]}(\alpha_i^4-3\alpha_i^2)\indicator{|\alpha_i|\le \tau^{1/4}})^d\indicator{E_1}$.  Then, 
	\begin{align}
	\Exp\left[Q(\alpha_S)^{d-1}\Eone\right] = \Exp\left[|Z|^{d-1}\indicator{Z\ge {\gamma'}\sqrt{nd}}\right]\label{eqn:15}
	\end{align}
We have that $Z_i$ is a sub-exponential variable with parameter $(\sqrt{41}, (4+o_{\tau}(1))\tau^{1/2})$ (by Lemma~\ref{lem:bernstein_moment}). Then by Lemma~\ref{lem:sub-exp-addition}, we have that $\sum_i Z_i$ is sub-exponential with parameters $(\sqrt{41n}, (4+o_{\tau}(1))\tau^{1/2})$. Moreover, we have $Z$ has mean $o_d(n)$ since $n \ge d\log d$. 
	
	Then, we use Lemma~\ref{lem:sub_exponential_moments} to control the moments the sub-exponential random variable $Z\cdot \indicator{|Z|\ge {\gamma'} \sqrt{nd}}$. Taking $s = {\gamma'} \sqrt{nd}$ and $\nu = \sqrt{41n}$ and $b= (4+o_{\tau}(1))\tau^{1/2}$ in Lemma~\ref{lem:sub_exponential_moments}, we obtain that when when ${\gamma'} \ge \max\{\sqrt{41}, (1+\eta)(8+o_{\tau}(1))\sqrt{\tau d/n}\} = \max\{\sqrt{41}, (1+\eta)(8+o_{\tau}(1))\sqrt{K}\}$
	\begin{align}
		\Exp\left[|Z|^{d-1} \indicator{Z\ge {\gamma'} \sqrt{nd}}\right]\le o_d(n)/\epsilon^{d} + \frac{1}{(1-\epsilon)^d}\left((1+o(1))^dd^2e^{-{\gamma'}^2d/82} +1/\eta \cdot e^{-(1/8-o_{\tau}(1)){\gamma'}/\sqrt{K}}\right)\nonumber\mper
	\end{align}
	Take $\epsilon = 1/\log d$ and $\eta = 1/\log d$,  we obtain that 
	\begin{align}
	\Exp\left[|Z|^{d-1} \indicator{Z\ge {\gamma'} \sqrt{nd}}\right]\le (1+o(1))^d ({\gamma'} \sqrt{nd})^{d/2}\left(e^{-{\gamma'}^2d/82} + \cdot e^{-(1/8-o_{\tau}(1)){\gamma'}/\sqrt{K}}\right)\label{eqn:38}\mper
	\end{align}
Hence, combing equation~\eqref{eqn:11}, ~\eqref{eqn:15} and~\eqref{eqn:38}, we have, 	\begin{align}
			\Exp\left[h(\alpha)\indicator{F_0}\right] & \lesssim  \Vol(\sphere^{d-1}) \Exp\left[W(\alpha)  p_{g\mid \alpha}(0)\right]\tag{by equation~\eqref{def:Wa}} \\
			&\lesssim \frac{\pi^{d/2}}{(d/(2e))^{d/2}} \cdot (1+o(1))^d (30\pi n)^{-d/2}({\gamma'}^2 nd)^{d/2}\left(e^{-{\gamma'}^2d/82} + \cdot e^{-(1/8-o_{\tau}(1)){\gamma'}/\sqrt{K}}\right) \nonumber\\
			& \lesssim(1+o(1))^d \left(\frac{e{\gamma'}^2}{15}\right)^{d/2}\left(e^{-{\gamma'}^2d/82} + \cdot e^{-(1/8-o_{\tau}(1)){\gamma'}/\sqrt{K}}\right)\le 2^{-d/2}\mper \tag{Since ${\gamma'}$ is a sufficiently large constant that depends on the choice of $K$}
	\end{align}
\end{proof}

\subsection{Proof of Proposition~\ref{prop:f_k}}

In the proof of Proposition~\ref{prop:f_k}, we will have another division of the the coordinates of $\alpha$ into two subsets $S$ and $L$ according to the magnitude of $\alpha_i$'s. Throughout this subsection, let $C$ be a sufficiently large constant that depends on $\delta$. Let $S = \{i: |\alpha_i| \le \sqrt{C\log d}\}$ and $L = [n]\backslash S$. Thus by definition, since $n \ge d\log^2 d$, we have that event $F_k$ implies that $k\in L$. But we note the threshold $\sqrt{C\log d}$ here is smaller than the threshold $\tau = Kn/d$ for defining event $F_k$'s. The key property that we use here is that the set $L$ have cardinality at most $d$ as long as $E_0$ happens. This will allows us to bound the contribution of the coordinates in $L$ in a more informative way.

\begin{claim}\label{claim:sizeL}
	Let $S,L$ are defined as above with $C\ge 2/\delta$. Suppose $E_0$ happens, then we have $|L|\le \delta n/\log d$. 
\end{claim}
\begin{proof}
	Assume for the sake of contradiction that $|L| > \delta n/\log d$, then we can pick a subset $L'\subset L$ of size $\delta n/\log d$ and obtain that $\norm{\alpha_{L'}}^2 \ge |L'| C\log d= \delta d C \ge 2d$ which violates~\eqref{eqn:RIP}.
\end{proof}

For notational convenience, let $P = (2-O(\delta))\tau^{1/2}d = (2-O(\delta))\sqrt{Knd}$. Recall $\alpha_S$ is the restriction of vector $\alpha$ to the coordinate set $S$, and $Q(\cdot): \R^d \rightarrow \R$ is defined as: 
\begin{align}
Q(z) = \Norm{z}_4^4 - 3\Norm{z}^2 \nonumber\mper
\end{align}
Define $T_1(\alpha),T_2(\alpha)$ as\begin{align}
T_{1}(\alpha) & =  \exp\left(-\frac{(P-Q(\alpha_S))^2}{36\norm{\alpha_S}_4^4}\right)\nonumber\\
T_{2}(\alpha) &=  \indicator{P \le Q(\alpha_S)}\mper\label{eqn:def-T12}
\end{align}

Therefore, since $Q(\alpha_S) = \norm{\alpha_S}_4^4-3\norm{\alpha_S}^2$ is roughly a random variable with standard deviation $\Theta(\sqrt{n})$, for large enough constant $K$, we should think of $T_1(\alpha)$ and $T_2(\alpha)$ as exponentially small random variable.  The following Lemma is the key of this section, which says that when $F_0$ happens, the chance of $M$ being PSD is very small. 

\begin{lemma}\label{lem:prob_indicator} In the setting of Proposition~\ref{prop:f_k}, let $T_{1}(\alpha),T_2(\alpha)$ be defined as in~\eqref{eqn:def-T12}. Then,
	\begin{align}
	\Exp\left[\indicator{M\succeq 0} \indicator{E_2'}\mid g, \alpha \right] \indicator{F_k} \indicator{E_0}\le T_{1}(\alpha) + T_2(\alpha) \mper\nonumber
	\end{align}
	\Tnote{We might be able to get slightly smaller constant}
\end{lemma}

\begin{proof}\Tnote{This proof requires another round of proofread}
	We assume throughout this proof that $\alpha$ is a fixed vector and that $\alpha$ satisfies $F_k$ and $E_0$. 	For simplicity, define $\hat{\tau} = \sqrt{C\log d}$. 
	
	Let $q = (\alpha_1^3,\dots, \alpha_n^3)^{\top}\in \R^n$, and $\bar{q} = q/\|q\|$.	Let $z_1,\dots z_{d-1}$ be i.i.d Gaussian random variables in $\R^d$ with covriance $\N(0,\Id_n-\bar{q}\bar{q}^{\top})$. Let $Z$ denote the matrix with $z_i^{\top}$ as rows and let $v_j~(j\in [n])$ denotes the $j$-th column of $Z$. Then, by Claim~\ref{claim:remove_conditioning}, random variable $(M, \indicator{E_2'})\mid g = 0, \alpha)$ has the same joint distribution as $(M', \indicator{E_2''})$ defined below \Tnote{To add explanation},
	\begin{align}
	M' = \|\alpha\|_4^4\cdot \Id_{d-1} -3\sum_{i\in [n]} \alpha_i^2v_iv_i^{\top}, E_2'' = \Set{\forall i, \|v_i\|^2 \ge (1-\delta)d} \mper\nonumber
	\end{align}
	Then, it suffices to bound from above the quantity 	$	\Pr\left[M'\succeq 0\wedge E_2''\right]\mper$
	
	A technical complication here is that $v_1,\dots, v_n$ are not independent random variables. We get around this issue by the following ``coupling'' technique which introduces some additional randomness. Let $\beta\in \N(0,\Id_{d-1})$ be a random variable that is independent with all the $v_i$'s and  define $\tilde{B} = \beta\bar{q}^{\top}  + Z$. We can verify that $\tilde{B}$ contains independent entries with distribution $\N(0,1)$. Let $\tilde{b}_1,\dots, \tilde{b}_n$ be the columns of $\tilde{B}$. Let $\bar{v}_k = v_k/\norm{v_k}$.  We have that $M'\succeq 0$ implies that 
	\begin{align}
	\bar{v}_k^{\top}M'\bar{v}_k= \|\alpha\|_4^4 - 3\sum_{i\in [n]}\alpha_i^2 \inner{v_i,\bar{v}_i}^2 \ge 0 \mper\label{eqn:23}
	\end{align}
	We bound from above $\bar{v}_k^{\top}M'\bar{v}_k$ by decomposing it according to the subset $S,L$. 
	\begin{align}
	\bar{v}_k^{\top}M'\bar{v}_k & = \norm{\alpha_{S}}_4^4 -3 \sum_{i\in S}\alpha_i^2 \inner{v_i,\bar{v}_i}^2 + \norm{\alpha_L}_4^4 - 3\sum_{i\in L} \alpha_i^2 \inner{v_i,\bar{v}_i}^2 \nonumber\\
	&\le \underbrace{\norm{\alpha_{S}}_4^4 - 3\sum_{i\in S}\alpha_i^2 \inner{v_i,\bar{v}_i}^2}_{A_S} + \underbrace{\norm{\alpha_L}_4^4 - 3\alpha_k^2\inner{v_k,\bar{v}_k}^2}_{A_L} \label{eqn:24}
	\end{align}
	Here $A_S, A_L$ are defined above in equation~\eqref{eqn:24}. 
	By Claim~\ref{claim:sizeL}, we have $|L| \le \delta n/\log d$.  By~\eqref{eqn:RIP} again, we have that $E_0$ implies that $\norm{\alpha_L}^2\le (1+\delta)d$. 
		Then, we have $\norm{\alpha_L}_4^4 \le \max_i \alpha_i^2 \cdot \norm{\alpha_L}^2\le (1+\delta)d\max_i \alpha_i^2 = (1+\delta)d\alpha_k^2$. Moreover, when $E_2''$ happens, we have that $\alpha_k^2\inner{v_k,\bar{v}_k}^2 = \alpha_k^2\norm{\bar{v}_k}^2 \ge \alpha_k^2 (1-\delta)d $. Thus, 	\begin{align}
	A_L\cdot \indicator{E_2''} &= (\norm{\alpha_L}_4^4 - 3\alpha_k^2\inner{v_k,\bar{v}_k}^2)\indicator{E_2''}  \le (1+\delta)d\alpha_k^2 - 3\alpha_k^2(1-\delta)d \nonumber\\ & \le - (2-O(\delta))\alpha_k^2 d \le -(2-O(\delta))\sqrt{\tau}d\label{eqn:25}
	\end{align}
	Then combining equation~\eqref{eqn:23}, ~\eqref{eqn:24} and~\eqref{eqn:25},  we have that when $M'\succeq 0$ and $E_2''$ both happen, it holds that $A_S\ge (2-O(d))\tau^{1/2}d$. 
	
	\Tnote{Require edits below}
	
	Next we consider in more detail the random variable $A_S$. Define random variable $Y = \bar{v}_k^{\top} \tilde{B}_{-k}\diag((\alpha_{S})_{-k})$, where $\tilde{B}_{-k}$ is the $(d-1)\times (n-1)$ sub-matrix of $\tilde{B}$ that excludes the $k$-th column of $\tilde{B}$, and $(\alpha_{S})_{-k}$ is the $n-1$-dimensional vector that is obtained from removing the $k$-th entry of $\alpha_S$. Note that since $k\not\in S$, we have that $\norm{Y}^2 = \sum_{i\in S}\alpha_i^2 \inner{v_i,\bar{v}_i}^2$, and $A_S  = \norm{\alpha_S}_4^4 - 3\norm{Y}^2$. Thus our goal next is to understand the distribution of $Y$. 
	
	We observe that $Y\mid v_k$ is a Gaussian random variable since $\tilde{B}_k\mid v_k$ is a Gaussian random variable. Let $\ell_i\in \R^{n-1}$ denote the $i$-th row of the matrix $\tilde{B}$. Then $\ell_1,\dots, \ell_{d-1}$ are i.i.d Gaussian random variable with distribution $\N(0,\Id_{n})$. Now we consider $\ell_i\mid v_{k,i}$, where $v_{k,i}$ is the $i$-th coordinates of vector $v_k$.  Note that $v_{k,i} = z_{i,k} = ((\Id-\bar{q}\bar{q}^{\top})\ell_i)_k = \inner{e_k- \bar{q}_k\bar{q}, \ell_i}$ (here $z_{i,k}$ denotes the $k$-th coordinate of $z_i$ and $e_k$ is the $k$-th natural basis vector in $\R^n$). Thus, conditioning on $v_{k,i}$ imposes linear constraints on $\ell_i$. Let $s_k = e_k- \bar{q}_k\bar{q}\in \R^n$ and $\bar{s}_k  = s_k/\norm{s_k}$. Then $\ell_i \mid v_{k,i}$ has covariance $\Id_{n} - \bar{s}_k\bar{s}_{k}^{\top}$. 	Computing the mean of $\ell_i \mid v_{k,i}$  will be tedious but fortunately our bound will not be sensitive to it. Now we consider random variable  $\ell_i' = (\ell_i)_{-k}\diag((\alpha_{S})_{-k})\mid v_{k,i}$ in $\R^{n-1}$ where $(\ell_i)_{-k}$ denotes the restriction of $\ell_{i}$ to the subset $[n]\backslash \{k\}$. Note that $\ell_i'$ is precisely a row of  $\tilde{B}_{-k}\diag((\alpha_{S})_{-k})$. We have $\ell_i'$ is again a Gaussian random variable and its covariance matrix is $\Sigma = \diag((\alpha_{S})_{-k})\left(\Id_{n} - \bar{s}_k\bar{s}_{k}^{\top}\right)_{-k,-k}\diag((\alpha_{S})_{-k})$, where $\left(\Id_{n} - \bar{s}_k\bar{s}_{k}^{\top}\right)_{-k,-k}$ is the restriction of $\Id_{n} - \bar{s}_k\bar{s}_{k}^{\top}$ to the subsets $([n]\backslash \{k\})\times ([n]\backslash \{k\})$. Since $\ell_i$'s are independent with each other, we have $Y^{\top} \mid \bar{v}_k= [\ell_1',\dots, \ell_{d-1}']\cdot \bar{v}_k$  is Gaussian random variable with covariance $\|\bar{v}_k\|^2\Sigma = \Sigma$. 
	
	Next we apply Lemma~\ref{lem:norm_concentration} on random variable $Y^{\top}$ and obtain that $Y^{\top}$ has concentrated norm in the sense that for any choice of $v_k$, 
	\begin{align}
	\Pr\left[\norm{Y}^2 \ge \trace(\Sigma) -\norm{\Sigma} - 2\sqrt{\trace(\Sigma^2)t}\mid v_k\right]\ge 1-e^{-t} \mper\label{eqn:30}
	\end{align}
	
	Recall that $\Sigma = \diag((\alpha_{S})_{-k})\left(\Id_{n} - \bar{s}_k\bar{s}_{k}^{\top}\right)_{-k,-k}\diag((\alpha_{S})_{-k})$. Therefore, we have $\trace(\Sigma) = \sum_{i\in S}\alpha_i^2 - \norm{\bar{s}_k\odot (\alpha_S)_{-k}}^2\ge \sum_{i\in S}\alpha_i^2 - \norm{\alpha_S}_{\infty}^2\ge \sum_{i\in S}\alpha_i^2 - \hat{\tau}^{1/2}$. Moreover, we have that $\norm{\Sigma}\le \norm{\alpha_S}^2\le \hat{\tau}^{1/2}$ and $\trace(\Sigma^2)\le \norm{\alpha_S}_4^4$.  Plugging these into equation~\eqref{eqn:30}, and using the fact that $A_S  = \norm{\alpha_S}_4^4 - 3\norm{Y}^2$, we have that for any $t\ge 0$, 
	\begin{align}
	\Pr\left[A_S \ge \norm{\alpha_S}_4^4 - 3 \norm{\alpha_S}^2 + 6\hat{\tau}^{1/2} + 6\sqrt{\norm{\alpha_S}^4_4 t}\mid v_k\right]\le e^{-t} \mper \label{eqn:31}
	\end{align}
	
	For simplicity, let $P = (2-O(\delta))\tau^{1/2}d - 6\hat{\tau}^{1/2} = (2-O(\delta)-o(1))\tau^{1/2}d$ and $Q(\alpha) = \|\alpha_S\|_4^4 - 3\|\alpha_S\|^2$. Then we have that suppose $P > Q(\alpha)$, by taking $t = \frac{(P-Q(\alpha))^2}{36\norm{\alpha_S}^2}$ in equation~\eqref{eqn:31}, 
	\begin{align}
	\Pr\left[A_S \ge P\mid v_k\right]& \le \exp\left(-\frac{(P-Q(\alpha))^2}{36\norm{\alpha_S}_4^4}\right)	\end{align} 	Hence, we have 
	\begin{align}
	\Pr\left[M'\succeq 0 \wedge E_2''\right]\le \Exp\left[\Pr\left[A_S \ge P\mid v_k\right]\right] \le \left\{\begin{array}{cc}
	\exp\left(-\frac{(P-Q(\alpha))^2}{36\norm{\alpha_S}_4^4}\right) & \textup{if $P > Q(\alpha)$} \\
	1 & \textup{otherwise}
	\end{array}\right.\nonumber
	\end{align}
\end{proof} 
 
For notational convenience, again we define $T_3(\alpha)$ and $T_4(\alpha)$ as,
\begin{align}
T_3(\alpha)  &=\left(\frac{Q(\alpha_S)^2}{\sixmoments{\alpha_S}} \right)^{d/2} \nonumber\\
T_4(\alpha) &=\left(\frac{Q(\alpha_L)^2}{\sixmoments{\alpha_L}} \right)^{d/2}\mper\label{eqn:def_Ts}
\end{align}
Intuitively, $T_3$ and $T_4$ are the contribution of the small coordinates and large coordinates to the quantity $\Exp[\trace(M)]^d p_{g\mid \alpha}(0) \propto Q(\alpha)(\norm{\alpha}_6^6)^{-d/2}$, which is the central object that we are controlling. 

Using Lemma~\ref{lem:prob_indicator}, we show that bounding $\Exp\left[h(\alpha)\indicator{F_k}\right]$ reduces to bounding the second moments of $T_1(\alpha), \dots, T_{4}(\alpha)$. 
\begin{lemma} \label{lem:Ts}
	Let $T_1(\alpha),T_2(\alpha), T_3(\alpha), T_4(\alpha)$ be defined as above. In the setting of Proposition~\ref{prop:f_k}, we have, 
	\begin{align}
	\Exp\left[	h(\alpha)\indicator{F_k}\right] 
& \lesssim \pi^{d/2} (d/(2e))^{-d/2} \max\{\Exp[T_1(\alpha)^2]^{1/2},\Exp\left[T_2(\alpha)^2\right]^{1/2}\}\nonumber\\
& ~~\cdot  \max\left\{\Exp\left[T_3(\alpha)^2\right]^{1/2},\Exp\left[T_4(\alpha)^2\right]^{1/2}, (d/1500)^{d/2}\right\}\mper \nonumber
	\end{align}
\end{lemma}

\begin{proof}[Proof of Lemma~\ref{lem:Ts}] 	By AM-GM inequality (Claim~\ref{claim:amgm}) we obtain that 
	\begin{align}
	\det(M)\indicator{M\succeq 0}& \le (d-1)^{-(d-1)}|\trace(M)|^{d-1}\indicator{M\succeq 0}\mper \label{eqn:50}
	\end{align}
	It follows that, 
	\begin{align}
	W(\alpha)\indicator{F_k} 
	& \le \Exp\left[\frac{|\trace(M)|^{d-1}}{(d-1)^{d-1}}\indicator{M\succeq 0} \indicator{F_k} \mid g = 0, \alpha\right] \indicator{E_0} \tag{by equation~\eqref{eqn:50}} \nonumber \\
	& \le	\underbrace{\Exp\left[\left(\frac{|\trace(M)|}{d-1}\right)^{2(d-1)}\mid g = 0, \alpha\right]^{1/2}}_{:=T_5(\alpha)} \underbrace{\Exp\left[\indicator{M\succeq 0}\indicator{F_k}\mid g = 0, \alpha\right]^{1/2}\indicator{E_0}}_{:=T_6(\alpha)} \label{eqn:46}
	\end{align}
	Here in the last step we used Cauchy-Schwarz inequality. 
	\Tnote{This might be another place that we lose some constant }
		Let $T_5(\alpha)$ be defined as in equation~\eqref{eqn:46}, by Corollary~\ref{cor:tracepower}
	\begin{align}
	T_{5}(\alpha)^2\indicator{E_0}\indicator{E_2} \le (1+O(\delta))^{2d} \max\left\{Q(\alpha)^{2d}, (0.1d)^{2d}, \left(0.1pd^{-1/2}n^{1/2}\right)^{2d}\right\}\mper\label{eqn:52}
	\end{align}

	Then, combining  equation~\eqref{eqn:52} and~\eqref{eqn:41}, we obtain that 
	\begin{align}
	T_5(\alpha)	p_{g\mid \alpha}(0) \indicator{E_0}\indicator{E_2}
	& \le (1+O(\delta))^d (2\pi)^{-d/2} (\Norm{\alpha}_6^6)^{-d/2} \max\left\{Q(\alpha)^d, \left(0.1d^{1/2}n^{1/2}\right)^{d}\right\} \indicator{E_0}\nonumber\\
	& \le (1+O(\delta))^d (\pi)^{-d/2}\max\left\{\left(\frac{Q(\alpha_S)^2}{\sixmoments{\alpha_S}} \right)^{d/2},\left(\frac{Q(\alpha_L)^2}{\sixmoments{\alpha_L}} \right)^{d/2}, \left(d/1500\right)^{d/2}\right\} \indicator{E_0}\label{eqn:57}
	\end{align}
		Here the last inequality uses Cauchy-Schwarz inequality and Holder inequality (technically,  Lemma~\ref{lem:1} with $\eta = 1/2$). \Tnote{Could be tightened by choosing other $\eta$. }
	Using Lemma~\ref{lem:prob_indicator}, we obtain that 
	\begin{align}
		T_6(\alpha)
	\le \exp\left(-\frac{(P-Q(\alpha_S))^2}{36\norm{\alpha_S}_4^4}\right)  + \indicator{P \le Q(\alpha_S)}\mper\label{eqn:62}\end{align}

 Now we have that our final target can be bounded by 
	\begin{align}
	\Exp\left[	h(\alpha)\indicator{F_k}\right] &= \Exp\left[\Vol(\sphere^{d-1}) W(\alpha) p_{g\mid \alpha}(0)\indicator{F_k}\right] \nonumber\\
	&\lesssim \pi^{d/2} (d/(2e))^{d/2} \Exp\left[ T_6(\alpha)\Fk (T_5(\alpha) p_{g\mid \alpha}(0) \indicator{E_0}\indicator{E_2})\right] \tag{by equation~\eqref{eqn:46}}\\
	& \le  \pi^{d/2} (d/(2e))^{-d/2} \Exp\left[\max\{T_1(\alpha),T_2(\alpha)\}\cdot \max\left\{T_3(\alpha),T_4(\alpha), (d/1500)^{d/2}\right\}\indicator{E_0}\right] \tag{by equation~\eqref{eqn:62} and~\ref{eqn:57}}\\
	& \lesssim \pi^{d/2} (d/(2e))^{-d/2} \max\{\Exp[T_1(\alpha)^2\Ezero]^{1/2},\Exp\left[T_2(\alpha)^2\right]^{1/2}\}\nonumber\\
	& ~~\cdot  \max\left\{\Exp\left[T_3(\alpha)^2\right]^{1/2},\Exp\left[T_4(\alpha)\right]^{1/2}, (d/1500)^{d/2}\right\} \mper \tag{by Cauchy-Schwarz inequality}
	\end{align}
	\end{proof}
	
The next few Lemmas are devoted to controlling the second moments of $T_1(\alpha),\dots, T_{4}(\alpha)$. We start off with $T_{3}(\alpha)$. 
\begin{lemma}\label{lem:moment_small_entries}
In the setting of Lemma~\ref{lem:Ts}, we have, 
	\begin{align}
	\Exp\left[T_3(\alpha)^2\Ezero\right]^{1/2}
	& \le (1+O(\delta))^d \left(\frac{82d}{15e}\right)^{d/2}\mper\nonumber
	\end{align}
\end{lemma}

\begin{proof}
		Let $Z = \Norm{\alpha_S}_4^4 - \Norm{\alpha_S}^2 = \sum_{i=1}^{n} (\alpha_i^4-3\alpha_i^2)\indicator{|\alpha_i| \le \sqrt{C\log d}}$. Then $Z$ is a sub-exponential random variable with parameters $(\sqrt{41n}, 4C\log d)$ (by Lemma~\ref{lem:bernstein_moment}). Moreover, for sufficiently large $C$ (depending on the choice of $\delta$) we have $|\Exp\left[Z\right]|\le \delta$. Therefore, by Lemma~\ref{lem:sub-exp-before_truncation-moments}, choosing $\epsilon = O(\delta)$, we have 
	\begin{align}
	 \Exp\left[Q(\alpha_S)^{2d}\right]=\Exp\left[|Z|^{2d}\right] &\le (1+O(\delta))^{2d}\cdot ((\sqrt{41n})^{2d} (2d)!! + (2C\log d)^{2d} (2d)!) \nonumber\\
	& \le (1+O(\delta))^d(\sqrt{41n})^{2d} (2d)!! \tag{\FIXME{?}Since $\delta n\ge 2d\log d$}\nonumber\\
	& \le (1+O(\delta))^d \left(82nd/e\right)^{d}\nonumber\mper
	\end{align}
	Moreover, when $E_0$ happens, we have $\Norm{\alpha_S}_6^6\ge 15(1-\delta)n$. Hence,
	\begin{align}
	\Exp\left[T_3(\alpha)^2\Ezero\right] = \Exp\left[Q(\alpha_S)^{2d} (\Norm{\alpha_S}_6^6)^{-d}\indicator{E_0}\right] & \le 	(1+O(\delta))^d \left(82nd/e\right)^{d}\cdot  (15(1-\delta)n)^{-d}\nonumber\\
	& \le (1+O(\delta))^d \left(\frac{82d}{15e}\right)^{d}\mper \nonumber
	\end{align}
\end{proof}

Next we control the second moments of $T_4(\alpha)$. 
\begin{lemma}\label{lem:moment_large_entries}
	In the setting of Lemma~\ref{lem:Ts}, we have
	\begin{align}
	\frac{Q(\alpha_L)^2}{\sixmoments{\alpha_L}} \indicator{E_0}\le (1+\delta) d\nonumber\mper
	\end{align}
	As a direct consequence, we have 
	\begin{align}
	\Exp\left[T_4(\alpha)^2\indicator{E_0}\right]^{1/2} \le (1+\delta)^d d^{d/2}\mper\nonumber
	\end{align}
\end{lemma}

\begin{proof}
	We have 
	\begin{align}
	\left(\norm{\alpha_L}^4-3\norm{\alpha_L}^2\right)^2 \indicator{E_0}
	& \le \left(\sum_{i\in L} \alpha_i^6\right)  \left(\sum_{i\in L}(\alpha_i-3/\alpha_i)^2\right) \tag{By Cauchy-Schwarz Inequality}\\
	&\le \Norm{\alpha_L}_6^6\Norm{\alpha_L}^2\indicator{E_0}\tag{since for $i\in L$, $(\alpha_i-3/\alpha_i)^2\le \alpha_i^2$}\\
	& \le \sixmoments{\alpha_L} (1+\delta) d\indicator{E_0}\mper\tag{since when $E_0$ happens, $\norm{\alpha}\le (1+\delta)d$}\end{align}
				\end{proof}

The next two Lemmas control the second moments of $T_1(\alpha)$ and $T_2(\alpha)$. 

\begin{lemma}\label{lem:2}
	In the setting of Lemma~\ref{lem:Ts},  we have  
	\begin{align}
	\Exp\left[T_2(\alpha)^2\right] \le (0.1)^d\mper
	\end{align}
\end{lemma}
\begin{proof}
Since  $Q(\alpha_S)$ is a sub-exponential random variable with parameters $(\sqrt{41n}, 4C\log d)$ (by Lemma~\ref{lem:bernstein_moment}), we have that (by Lemma~\ref{lem:sub-exp-property})
\begin{align}
\Pr\left[Q(\alpha)\ge P\right] & \le \exp((2-O(\delta))Kd/82) + \exp(-\sqrt{Knd}/(8C\log d))\nonumber \\
& \lesssim \exp((2-O(\delta))Kd/82) \le (0.01)^d \mper\tag{since $K \ge 2000$ and $n\ge \beta d\log^2 d$ for sufficiently large $\beta$ (that depends on $\delta$)}
\end{align}
\end{proof}

\begin{lemma}\label{lem:3}
	In the setting of Lemma~\ref{lem:Ts},  we have  
	\begin{align}
	\Exp\left[T_1(\alpha)^2\Eone\right] \le (0.1)^d
	\end{align}
	\end{lemma}

\begin{proof}
	Similarly to the proof of Lemma~\ref{lem:2}, we have that for $K \ge 400$, it holds that $\Pr\left[Q(\alpha)\ge P/2\right] \le (0.1)^d$. Then it follows that 
	\begin{align}
	\Exp\left[T_1(\alpha)^2\Ezero\right] & \le \Exp\left[1\mid Q(\alpha)\ge P/2\right]\Pr\left[Q(\alpha)\ge P/2\right]  \nonumber\\
	& + \Exp\left[\exp(-\frac{P^2}{144\fourmoments{\alpha_S}})\Ezero\mid Q(\alpha) < P/2\right]\Pr\left[Q(\alpha)< P/2\right] \nonumber \\
	& \le (0.1)^d + \exp(-Kd/432) \lesssim (0.01)^d\mper  \tag{for $K \ge 2\cdot10^3$.}
	\end{align}
\end{proof}

Finally using the four Lemmas above and Lemma~\ref{lem:Ts}, we can prove Proposition~\ref{prop:f_k}.

\begin{proof}[Proof of Proposition~\ref{prop:f_k}]
	Using Lemma~\ref{lem:Ts}, Lemma~\ref{lem:moment_small_entries},~\ref{lem:moment_large_entries}, \ref{lem:2},\ref{lem:3}, we have that 
	\begin{align}
	\Exp\left[	h(\alpha)\indicator{F_k}\right] 
	& \lesssim \pi^{d/2} (d/(2e))^{d/2} \max\{\Exp[T_1(\alpha)^2]^{1/2},\Exp\left[T_2(\alpha)^2\right]^{1/2}\}\nonumber\\
	& ~~\cdot  \max\left\{\Exp\left[T_3(\alpha)^2\right]^{1/2},\Exp\left[T_4(\alpha)\right]^{1/2}, (d/1500)^{d/2}\right\} \nonumber \\
	& \lesssim \pi^{d/2} (d/(2e))^{-d/2}\cdot\left(\frac{82d}{15e}\right)^{d/2} (0.01)^{d/2}\le (0.3)^d\mper\nonumber\end{align}
\end{proof}

%% file: local_stripped.tex
\section{Local Maxima Near True Components}
\label{sec:local}
In this section we show that in the neighborhoods of the $2n$ true components, there are exactly $2n$ local maxima. 
Recall the set $L_2$ was defined to be the set of points $x$ that do not have large correlation with {\em any} of the components:
$$
L_2 := \{x\in \sphere^{d-1}: \forall i, \norm{P_x a_i}^2 \ge (1-\delta)d, \mathand |\inner{a_i,x}|^2\le \delta d\}\mper $$
We will show that the objective function (\ref{eq:obj}) has exactly $2n$ local maxima in the set $L_2^c$, and they are close to the normalized version of $\pm a_i$'s.

Let $\bar{a}_i = \frac{a_i}{\|a_i\|}$ be the normalized version of $a_i$ on the unit sphere. We prove the following slightly stronger version of Theorem~\ref{thm:local}. (Note that Theorem~\ref{thm:local} is a straightforward corollary of the Theorem below, since w.h.p, for every $i$, $\|a_i\|=\sqrt{d}\pm \widetilde{O}(1)$. )

\begin{theorem}Suppose  $(d\log d)/\delta^2 \le n \le d^2/\log^{O(1)} d$. Then, with high probability over the choice $a_1,\dots, a_n$, 
	\begin{align*}
	|\cM_f \cap L_1\cap  L_2^c| = 2n
	\end{align*}
Moreover, each of the point in $\cM_f\cap L\cap L_2^c$ is $\tilde{O}(\sqrt{n/d^3})$-close to one of $\pm \bar{a}_1,\dots \pm \bar{a}_n$. 
\end{theorem}

Towards proving the Theorem, we first show that all the local minima in the set $L_2^c$ must have at least $0.99$ correlation with one of $\pm \bar{a}_i$ (see Lemma~\ref{lem:highcor}). Next we show in each of the regions $\inner{x,\bar{a}_i} \ge 0.99$, the objective function is {\em strongly convex} with a unique local maximum (see Lemma~\ref{lem:unique}).

\begin{lemma}\label{lem:highcor}
In the setting of Theorem~\ref{thm:local}, for any local maximum $x\in L_2$, there exists a component $a \in \{\pm \bar{a}_i\}$ such that $\inner{x,a} \ge 0.99$.
\end{lemma}

We can partition points in $L_2$ according to their correlations with the components. We say its largest correlation is the maximum of $|\inner{\bar{a}_i,x}|$, and second largest correlation is the second largest among $|\inner{\bar{a}_i,x}|$. For any point $x$ in $L_2$, it is in one of the three types 1) the largest correlation is at least $0.99$; 2) the largest correlation is at least $\sqrt{2}$ larger than the second largest correlation; 3) the largest and second largest correlations are within a factor of $\sqrt{2}$. Intuitively, points in case 1 are what we want for the Lemma, points in case 2 will have a nonzero gradient, points in case 3 will not have a negative semidefinite Hessian (hence points in cases 2 and 3 cannot be local maxima). We formalize this intuition in the following proof:

\begin{proof}[Proof of Lemma~\ref{lem:highcor}]
Let $c_i = \inner{x, \bar{a}_i}$ , without loss of generality, we can rearrange the $a_i$'s so that 
$$
|c_1|\ge |c_2| \ge \cdots \ge |c_n|.
$$
Let $k$ be the index such that $|c_k| \ge \frac{\log d}{\sqrt{d}}$ and $|c_{k+1}| < \frac{\log d}{\sqrt{d}}$. We will call components $1,2,...,k$ the large components, and the rest the small components. We assume the following events happen simultaneously (note that, with high probability, all of them are true):

\begin{enumerate} 
\item $\{a_i\}$'s satisfy the $(d/\Delta\log n,0.01)$-RIP property for some universal constant $\Delta$. As a direct consequence, $k \ll d/\log d$ and $\sum_{i=1}^k c_i^2 \le (1.01)^2$.).  (See Definition~\ref{def:rip})
\item For all $i$, we have $\|a_i\| = \sqrt{d}(1\pm o(1))$.
\item For all $i\neq j$, $|\inner{a_i,a_j}| \le \sqrt{d}\log d$. As a consequence, we have $|\inner{\bar{a}_i,\bar{a}_j}| \le (1+o(1))\frac{\log d}{\sqrt{d}}$.
\item Concentration Lemmas~\ref{lem:degree4},~\ref{lem:degree3} and \ref{lem:degree2} hold.
\end{enumerate}

We aim to show $|c_1|\ge 0.99$. First, if $0.99 > |c_1| > \sqrt{2}|c_2|$, we will show the point has nonzero $\grad f(x)$, and hence cannot be a local maximum.

\begin{claim} \label{clm:powerupdate}
If $0.99 > |c_1| > \sqrt{2}|c_2|$, then $\frac{|\|P_{\bar{a}_1^\perp} \nabla f(x)\|}{|\inner{\nabla f(x),\bar{a}_1}|} < \frac{3}{4}\cdot  \frac{\|\Pi_{\bar{a}_1^\perp} x\|}{|\inner{x,\bar{a}_1}|} + \frac{O(\sqrt{nd}\log^4 d)}{|c_1|^3d^2}$, in particular when $|c_1|\le 0.99$ we know $\grad f(x)\ne 0$. 
\end{claim}

\begin{proof}
Let $x' = \nabla f(x)$, we know the gradient $\grad f(x) = 0$ if and only if $x'$ is a multiple of $x$. Therefore we shall show $x'$ is not a multiple of $x$ by showing $\frac{|\|\Pi_{\bar{a}_1^\perp} x'\|}{|\inner{x',\bar{a}_1}|} > \frac{\|\Pi_{\bar{a}_1^\perp} x\|}{|\inner{x,\bar{a}_1}|}$.
Intuitively, after one step of gradient ascent/power iterations, the point should become more correlated with $\pm a_1$.

Since $x' = \sum_{i=1}^n 4\inner{x,a_i}^3a_i$, we can write $x' = 4\inner{a_1,x}^3 a_1 + x'_L+x'_S$, where $x'_L = \sum_{i=2}^k 4\inner{x,a_i}^3a_i$ and $x'_S = \sum_{i=k+1}^n 4\inner{x,a_i}^3a_i$. The first term has norm $(c_1^3)(1\pm o(1))d^2$ and is purely in the direction $\bar{a}_1$. We just need to show the other terms are small. First, we bound $\|x'_S\|$ by concentration inequality Lemma~\ref{lem:degree3}: for any vector $x$ we know
\begin{equation}
\|\sum_{i=k+1}^n 4\inner{x,a_i}^3a_i\| \le O(n + \sqrt{nd} \log^4 d) = o((|c_1|^3)d^2).\label{eq:boundx'_S}
\end{equation}

On the other hand, we have
\begin{align}
\inner{x'_L,\bar{a}_1} & = \sum_{i=2}^k 4\inner{x,a_i}^3 \inner{a_i,\bar{a}_1} \nonumber \\
& \le \sum_{i=2}^k 4|c_i|^3 \|a_i\|^3 \log d && \mbox{$|\inner{a_i,a_j}| \le \sqrt{d}\log d$ and $\|a_i\| = (1\pm o(1))\sqrt{d}$} \nonumber
\\ & \le 4(1+o(1))d^{3/2}\log d \cdot |c_2|\sum_{i=2}^k |c_i|^2  && \mbox{$|c_2|$ is the largest} \nonumber\\
& \le O(d^{3/2}|c_2|\log d) = o(d^2). \label{eq:boundx'_Lcor}
\end{align}
Therefore $x'_L$ does not have large correlation with $\bar{a}_1$. We can also bound the norm of $\Pi_{\bar{a}_1^\perp} x'_L$: by RIP condition we know $\bar{a}_1,...,\bar{a_k}$ form an almost orthonormal basis, therefore

$$
\|\Pi_{\bar{a}_1^\perp} x'_L\|^2 \le 1.01^2\sum_{i=2}^k 16|c_i|^6. 
$$

We know $c_i = \inner{c_1\bar{a}_1, \bar{a}_i} + \inner{\Pi_{\bar{a}_1^\perp} x, \bar{a}_i}$, by Claim~\ref{clm:6moment} we know that $\|a+b\|_6^6 \le 1.01\|a\|_6^6 + O(\|b\|_6^6)$, therefore

\begin{align*}
\sum_{i=2}^k |c_i|^6 & \le 1.01 \sum_{i=2}^k \inner{\Pi_{\bar{a}_1^\perp} x, \bar{a}_i}^6 + O(\sum_{i=2}^k \inner{c_1\bar{a}_1, \bar{a}_i}^6) \\
& \le 1.01 \max_{i=2}^k \inner{\Pi_{\bar{a}_1^\perp} x, \bar{a}_i}^4 \sum_{i=2}^k \inner{\Pi_{\bar{a}_1^\perp} x, \bar{a}_i}^2 + O(|c_1|^6\log^6 d/d^2) \tag{since $|\inner{\bar{a_i},\bar{a_j}}| \le \frac{(1+o(1))\log d}{\sqrt{d}}$} \\
& \le 1.01^3(1+o(1))c_2^2\|\Pi_{\bar{a}_1^\perp} x\|^2 + O(|c_1|^6\log^6 d/d^2),
\end{align*}

where the last inequality follows from RIP condition as it implies $\sum_{i=2}^k \inner{\Pi_{\bar{a}_1^\perp} x, \bar{a}_i}^2 \le 1.01\|\Pi_{\bar{a}_1^\perp} x\|^2$.

Combining these we know 
\begin{equation}
\|\Pi_{\bar{a}_1^\perp} x'_L\| \le 4*1.01^{2.5} (1+o(1))c_2^2 \|\Pi_{\bar{a}_1^\perp} x\| + O(|c_1|^6\log^6 d/d^2). \label{eq:boundx'_L}
\end{equation}

Therefore
\begin{align*}
\frac{|\|\Pi_{\bar{a}_1^\perp} \nabla f(x)\|}{|\inner{\nabla f(x),\bar{a}_1}|} &\le \frac{\|x'_L\| + \|\Pi_{\bar{a}_1^\perp} x'_S\|}{4(1-o(1))|c_1|^3d^2 - |\inner{x'_S,\bar{a}_1}| - \|x'_S\|} \\
& \le \frac{\|x'_L\| + \|\Pi_{\bar{a}_1^\perp} x'_S\|}{4(1-o(1))|c_1|^3d^2}  \tag{by eqn. \eqref{eq:boundx'_Lcor} and~\eqref{eq:boundx'_S})} \\
& \le \frac{\sqrt{nd}\log^4 d + 4*1.01^{2.5}(1+o(1))c_2^2 \|\Pi_{\bar{a}_1^\perp} x\| + O(|c_1|^6\log^4 d/d^2)}{4(1-o(1))|c_1|^3d^2} \tag{by eqn. \eqref{eq:boundx'_S} and~\eqref{eq:boundx'_L})}\\
& \le \frac{3}{4}\cdot  \frac{\|\Pi_{\bar{a}_1^\perp} x\|}{|\inner{x,\bar{a}_1}|} + \frac{O(\sqrt{nd}\log^4 d)}{|c_1|^3d^2} \tag{since $c_1^2 \ge 2c_2^2$}
\end{align*}

\end{proof}

Next we show if $|c_1| \le \sqrt{2}|c_2|$, then the Hessian $\hessian f(x)$ is not negative semidefinite, hence again $x$ cannot be a local maximum

\begin{claim}
If  $|c_1| \le \sqrt{2}|c_2|$, we have $\sigma_{max}(\hessian f(x)) > 0$. 
\end{claim}

\begin{proof}
Consider the space spanned by $\bar{a}_1$ and $\bar{a_2}$, this is a two dimensional subspace so there is at least one direction that is orthogonal to $x$. Let $u$ such a direction ($u \in \mbox{span}(\bar{a}_1, \bar{a_2}), \inner{u,x} = 0, \|u\| = 1$). By Claim~\ref{claim:grad-hessian}, we know the Hessian is equal to 

$$\hessian f(x)  = 3 \sum_{i=1}^{n}\inner{a_i,x}^2 P_xa_ia_i^{\top}P_x - \left(\sum_{i=1}^{n} \inner{a_i,x}^4\right) P_x.$$

Clearly, the first term is PSD and the second term is negative. We will consider $u^\top \hessian f(x) u$, and break both terms in the Hessian into the sum of large and small components. For the large components, we know $(c_1^2 \|a_1\|^2 a_1a_1^{\top} + c_2^2\|a_2\|^2 a_2a_2^{\top})$ is a matrix that has smallest singular value at least $c_2^2d^2 (1-o(1))$ in subspace $\mbox{span}(\bar{a}_1,\bar{a_2})$ because $a_1$ and $a_2$ are almost orthogonal. Also, $u$ is orthogonal to $x$ so $P_x u = u$, therefore for large components
\begin{align*}
u^\top (3 \sum_{i=1}^{k}\inner{a_i,x}^2 P_xa_ia_i^{\top}P_x) u & \ge 3u^\top (c_1^2 \|a_1\|^2 P_xa_1a_1^{\top}P_x + c_2^2\|a_2\|^2 P_xa_2a_2^{\top}P_x) u 
\\ & \ge 3c_2^2 d^2 (1-o(1)).
\end{align*}

On the other hand, for the second term, the contribution from large components is smaller

$$
u^\top (\sum_{i=1}^{k} \inner{a_i,x}^4) P_x u = \sum_{i=1}^{k} \inner{a_i,x}^4 \le (1+o(1))d|c_1|^2\sum_{i=1}^k \inner{a_i,x}^2 \le (1.03+o(1))c_1^2 d^2.
$$

For the small components, we only count their contributions to the second term (because the first term is positive anyways), by concentration inequality Lemma~\ref{lem:degree4}, we know with high probability for {\em all} $u$

$$
u^\top (\sum_{i=k+1}^{n} \inner{a_i,x}^4) P_x u = \sum_{i=k+1}^{n} \inner{a_i,x}^4 \le 5n \ll c_2^2 d^2.
$$
The last inequality is because $c_2^2 \ge c_1^2/2 \ge \delta^2(1-o(1))/2$ and $n \ll d^2\delta^2$. Therefore, combining all these terms we know
$$
u^\top (\hessian f(x)) u \ge 3c_2^2 d^2 (1-o(1)) - (1.03+o(1))c_1^2 d^2 - 5n \ge 0.5c_2^2d^2 > 0.
$$

Therefore the Hessian is not negative semidefinite, and the point $x$ could not have been a local maximum.
\end{proof}

The Lemma follows immediately from the two claims.

\end{proof}

\begin{lemma}\label{lem:unique}
Under the same condition as Theorem~\ref{thm:local}, for all vector $z$ in $\{\pm \bar{a}_i\}$, in the set $\{x: \inner{x, z} \ge 0.99\}$there is a unique local maximum that is $\tilde{O}(\sqrt{n/d^3})$-close to $z$.   
\end{lemma}

The fact that power method converges to a local maximum in this region is proven by Anandkumar et al.\cite{anandkumar2015learning}. We give the proof for completeness.

\begin{proof}
Without loss of generality we assume $\inner{x,\bar{a}_1} \ge 0.99$.

By Claim~\ref{clm:powerupdate}, we know for any $x$ in the set $\{x: \inner{x, \bar{a}_1} \ge 0.99\}$, $\frac{\nabla f(x)}{\|\nabla f(x)\|}$ is in the same set. Therefore by Schauder fixed point theorem\cite{wiki:fixedpoint} there is at least a critical point in this set. Again by Claim~\ref{clm:powerupdate} we know every critical point must satisfy $\frac{|\|\Pi_{\bar{a}_1^\perp} \nabla f(x)\|}{|\inner{\nabla f(x),\bar{a}_1}|} \le \frac{O(\sqrt{nd}\log^4 d)}{|c_1|^3d^2}$. Therefore we only need to prove there is a unique critical point and it is also a local maximum. We do this by showing the Hessian $\hessian f(x)$ is always negative semidefinite in this set.

Again let $c_i = \inner{x, \bar{a}_i}$ and rearrange the $a_i$'s so that 
$$
|c_1|\ge |c_2| \ge \cdots \ge |c_n|.
$$
Let $k$ be the index such that $|c_k| \ge \frac{\log d}{\sqrt{d}}$ and $|c_{k+1}| < \frac{\log d}{\sqrt{d}}$. With high probability, we know $\{a_i\}$ satisfy the $(d/4,0.01)$-RIP property. Under RIP condition we know $k \ll d/\log d$ and $\sum_{i=1}^k c_i^2 \le (1+\zeta)^2$. We also condition on event that for all $i$ $\|a_i\| = \sqrt{d}(1\pm o(1))$ and for all $i,j$ $|\inner{a_i,a_j}| \le \sqrt{d}\log d$ which happens with high probability when $n \ge d\log d$.

Now consider the Hessian 
$$
\hessian f(x)  = 3 \sum_{i=1}^{n}\inner{a_i,x}^2 P_xa_ia_i^{\top}P_x - \left(\sum_{i=1}^{n} \inner{a_i,x}^4\right) P_x.
$$

The second term is obviously larger than $c_1^4 (1-o(1))d^2 = (1-o(1))d^2$. Therefore we only need to show the spectral norm of the first term is much smaller. We can break the first term into 3 parts: the first component, components 2 to $k$, and components $k+1$ to $n$. For the first component, we know in the set $\|P_xa_1\| \le (1+o(1))\sqrt{0.02d}$, so 
\begin{equation}
\|\inner{a_1,x}^2 P_xa_1a_1^{\top}P_x\| \le (1+o(1)) \cdot 0.02d^2.\label{eqn:bound1}
\end{equation}

For the second component, 
\begin{equation}
\|\sum_{i=2}^{k}\inner{a_i,x}^2 P_xa_ia_i^{\top}P_x\| \le (1+o(1))|c_2^2|d^2\|\sum_{i=2}^k \bar{a}_i\bar{a}_i^\top\| \le 0.03d^2. \label{eqn:bound2}
\end{equation}
Here the last inequality used the fact that $A$ is RIP and therefore $\sum_{i=2}^k \bar{a}_i\bar{a}_i^\top$ cannot have large eigenvalue. Finally for the third component we can bound it directly by concentration inequality Lemma~\ref{lem:degree2}, 

\begin{equation}
\|\sum_{i=k+1}^{n}\inner{a_i,x}^2 P_xa_ia_i^{\top}P_x\| \le O(n + \sqrt{nd}\log^4 d) \ll d^2.\label{eqn:bound3}
\end{equation}

Taking the sum of these three equations (\ref{eqn:bound1}-\ref{eqn:bound3}), we know $\hessian f(x)$ is always negative semidefinite in the set. Therefore the set can only have a unique critical point which is also a local maximum.
\end{proof}

%% file: manifold_opt_stripped.tex
\section{Brief introduction to manifold optimization}\label{sec:manifold}

Let $\cM$ be a Riemannian  manifold. For every $x\in \cM$, let $\tangent_x \cM$ be the tangent space to $\cM$ at $x$, and $P_x$ denotes the projection to the tangent space $\tangent_x \cM$. Let  $\grad f(x)\in \tangent_x \cM$ be the gradient of $f$ at $x$ on $\cM$ and $\hessian f(x)$ be the Riemannian Hessian. Note that $\hessian f(x)$ is a linear mapping from $\tangent_x \cM$ onto itself. 

In this paper, we work with $\cM =\sphere^{d-1}$, and we view it as a submanifold of $\R^d$. We also view $f$ as the restriction of a smooth function $\bar{f}$ to the manifold $\cM$. In this case, we have that $\tangent_x \cM= \{z\in \R^d: z^{\top} x = 0\}$,  and $P_x = \Id - xx^{\top}$.  The gradient can be computed by,  \begin{align}
\grad f(x) = P_x \nabla\bar{f}(x)\mcom\nonumber
\end{align}
where $\nabla$ is the usual gradient in the ambient space $\R^d$. Moreover, the Riemannian Hessian is defined as, 
\begin{align}
\forall \xi\in \tangent_x\cM, \quad \hessian f(x)[\xi]  & = P_x (\Diff\grad f(x) [\xi]) \nonumber\\
& = P_x \nabla^2 \bar{f}(x) \xi - (x^{\top}\nabla \bar{f}(x)) \xi \mper\nonumber
\end{align}

Here we refer the readers to the book~\cite{absil2007optimization} for the derivation of these formulae and the exact definition of gradient and Hessian on the manifolds.\footnote{For example, the gradient is defined in~\cite[Section3.6, Equation (3.31)]{absil2007optimization}, and the Hessian is defined in\cite[Section 5.5, Definition 5.5.1]{absil2007optimization}. ~\cite[Example 5.4.1]{absil2007optimization} gives the Riemannian connection of the sphere $\sphere^{d-1}$ which can be used to compute the Hessian.}

%% file: toolbox_stripped.tex
\section{Toolbox}
In this section we collect most of the probabilistic statements together. Most of them follow from basic (but sometimes tedious) calculation and calculus. We provide the full proofs for completeness. 
\subsection{Sub-exponential random variables}
In this subsection we give the formal definition of sub-exponential random variables, which is used heavily in our analysis. We also summarizes it's properties. 
\begin{definition}[{c.f.\cite[Definition 2.2]{wainwrightBasic}}]\label{def:sub-exponential}
	A random variable $X$ with mean $\mu = \Exp\left[X\right]$ is sub-exponential if there are non-negative parameters $(\nu,b)$ such that 
	\begin{align}
	\Exp\left[e^{\lambda (X-\mu)}\right] \le e^{\frac{\nu^2\lambda^2}{2}}, \quad \forall \lambda\in \R,  |\lambda| < 1/b\mper\nonumber
	\end{align}
\end{definition}
The following lemma gives a sufficient condition for $X$ being a sub-exponential random variable. 
\begin{lemma}\label{lem:conversion}
	Suppose random variable $X$ satisfies that for any $t\ge 0$, 
	\begin{align}
	\Pr\left[|X-\mu| \ge p\sqrt{t} + qt\right] \le 2e^{-t}\mper\nonumber
	\end{align}
	Then $X$ is a sub-exponential random variable with parameter $(\nu,b)$ satisfying $\nu \lesssim p$ and $b\lesssim q$. 
\end{lemma}

The following Lemma gives the reverse direction of Lemma~\ref{lem:conversion} which controls the tail of a sub-exponential random variable. 
\begin{lemma}[{\cite[Proposition 2.2]{wainwrightBasic}}]\label{lem:sub-exp-property}
	Suppose $X$ is a sub-exponential random variable with mean 0 and parameters $(\nu,b)$. Then, 
	\begin{align}
	\Pr\left[X\ge x\right]\le \max\{e^{-x^2/(2\nu^2)}, e^{-x/(2b)}\} \le e^{-x^2/(2\nu^2)}+e^{-x/(2b)}\mper \nonumber
	\end{align}
\end{lemma}
A summation of sub-exponential random variables remain sub-exponential (with different parameters). 
\begin{lemma}[c.f.~\cite{wainwrightBasic}]\label{lem:sub-exp-addition}
	Suppose independent random variable $X_1,\dots,X_n$ are sub-exponential variables with parameter $(\nu_1,b_1),\dots, (\nu_n,b_n)$ respectively, then $X_1+\dots+X_n$ is a sub-exponential random variable with parameter $(\mu^*,b^*)$ where $\nu^* = \sqrt{\sum_{k\in [n]} \nu_k^2}$ and $b^* = \max_k b_k$.  
\end{lemma}
The following Lemma uses the Lemma above to prove certain sum of powers of Gaussian random variables are sub-exponential random variables. 
\begin{lemma}\label{lem:bernstein_moment}
	Let $x\sim \N(0,1)$ and  $z = (x^4 -3x^2)\indicator{|x|\le \tau^{1/4}}$ where $\tau > C$ for a sufficiently large constant $C$. Then we have that $Z$ is a sub-exponential random variable with parameter $(\nu,b) $ satisfying $\nu = \sqrt{41}$ and $b=  (4+o_{\tau}(1))\sqrt{\tau}$. 
\end{lemma}
\begin{proof}
	We verify that $Z$ satisfies the Bernstein condition 
	\begin{align}
	\Exp\left[|z|^k\right]\le \frac{1}{2}k! \nu^2 b^{k-2} \quad \forall k \ge 2
	\end{align}
	For $k = 2$, we have that $\Exp\left[z^2\right] \le  \Exp\left[x^{8}-6x^6+9x^4\right] = 41 = \frac 1 2 \cdot 2! \cdot \nu^2$.  For any $k$ with $(1-o_{\tau}(1))\frac{e\sqrt{\tau}}{4}\ge k > 3$, we have 
	\begin{align}
	\Exp\left[|z|^k\right]	& \le \Exp\left[x^{4k}\indicator{\sqrt{3}\le |x|\le \tau^{1/4}}\right] + \left(\frac{9}{4}\right)^k \tag{since $|z|\le \frac{9}{4}$ when $|x|\le 3$} \\
	& = (4k-1)!! + (9/4)^k  \le 2^{2k} (2k)! + (9/4)^k \nonumber\\
	& \le 2^{2k} ((2k+1)/e)^{2k+1}+ (9/4)^k\tag{by Claim~\ref{claim:factorial}}\\
	& \le \frac{1}{2}(1+o_{\tau}(1)) k! (4\sqrt{\tau})^{k-2}\tag{since $(1-o_{\tau}(1))\frac{e\sqrt{\tau}}{4}\ge k$ adn $\tau$ is large enough}\\
	& \le \frac{1}{2}k! b^{k-2}
	\end{align}
	On the other hand, when $k\ge (1-o_{\tau}(1))\frac{e\sqrt{\tau}}{4}$, we have 
	\begin{align}
	\Exp\left[|z|^k\right] &\le \tau^k + \left(\frac{9}{4}\right)^k\tag{since $|z|\lesssim \tau^{1/4}$ a.s}\\
	& \le \left(\frac{4k\sqrt{\tau}}{e}(1+o_{\tau}(1))\right)^k \nonumber\\
	& \le \frac{1}{2}e(k/e)^k b^{k-2} \tag{since $k\gtrsim \sqrt{\tau}$ is large enough and $b =4 (1+o_{\tau}(1))\sqrt{\tau}$}\\
	& \le \frac{1}{2}k!b^{k-2}\nonumber
	\end{align}
	
\end{proof}

\subsection{Norm of Gaussian random variables}
The following Lemma shows that the norm of a Gaussian random variable has an sub-exponential upper tail and sub-Gaussian lower tail. 
\begin{lemma}[{\cite[Proposition 1.1]{hsu2012tail}}]\label{lem:quadratic_Gaussian}
	Let $A\in \R^{n\times n}$ be a matrix and $\Sigma = A^{\top}A$. Let $x\sim \N(0,\Id_n)$.  
	Then for all $t \ge 0$, 
	\begin{align}
&	\Pr\left[\|Ax\|^2 - \trace(\Sigma) \ge 2\sqrt{\trace(\Sigma^2)t} + 2\|\Sigma\|t \right] \le e^{-t}\nonumber \\
&		\Pr\left[\|Ax\|^2 - \trace(\Sigma) \le - 2\sqrt{\trace(\Sigma^2)t} \right] \le e^{-t} \mper\nonumber
	\end{align}
	Moreover, $\|Ax\|^2$ is a sub-exponential random variable with parameter $(\nu,b)$ with $\nu \lesssim \sqrt{\trace(\Sigma^2)}$ and $b\lesssim \|\Sigma\|$.
	
	\FIXME{~\cite[Lemma 1]{laurent2000} deosn't seem to be optimal in terms of constant. So this lemma could be improved potentially}
\end{lemma}

We note that this is a slight strengthen of{~\cite[Proposition 1.1]{hsu2012tail}} with the lower tail. The strengthen is simple. The key of ~\cite[Proposition 1.1]{hsu2012tail} is to work in the eigenspace of $\Sigma$ and apply~\cite[Lemma 1]{laurent2000}. The lower tail can be bounded exactly the same by using the lower tail guarantee of~\cite[Lemma 1]{laurent2000}. The second statement follows simply from Lemma~\ref{lem:conversion}. 

The following Lemma is a simple helper Lemma that says that the norm of a product of a Gaussian matrix with a vector  has a sub-Gaussian lower tail. 
\begin{lemma}\label{lem:norm_concentration}
	Let $v\in \R^m$ be a fixed vector and $z_1,\dots, z_m\in \R^{n}$ be independent Gaussian random with mean $\mu\in \R^n$ and covariance $\Sigma$. Then $X = [z_1,\dots, z_m]v$ is a Gaussian random variable with mean $(v^{\top}\allones)\mu$ and covariance $\|v\|^2 \Sigma$. 
	
	Moreover, we have that for any $t \ge 0$, 
	\begin{align}
	\Pr\left[\norm{X}^2 \ge \trace(\Sigma) -\norm{\Sigma} - 2\sqrt{\trace(\Sigma)t}\right]\ge 1-e^{-t} \mper\label{eqn:26}
	\end{align}
\end{lemma}

\begin{proof}
	The mean and variance of $X$ can be computed straightforwardly. Towards establishing~\eqref{eqn:26}, let $\bar{\mu} = \mu/\norm{\mu}$, and we define $X' = (\Id - \bar{\mu}\bar{\mu}^{\top})X$. Thus we have $\norm{X}\ge \norm{X'}$ and that $X'$ is Gaussian random variable with mean 0 and covariance $\Sigma' = (\Id - \bar{\mu}\bar{\mu}^{\top})\Sigma(\Id - \bar{\mu}\bar{\mu}^{\top})$. Then by Lemma~\ref{lem:quadratic_Gaussian}, we have that 
	\begin{align}
	\Pr\left[\norm{X}^2\ge \trace(\Sigma') - 2\sqrt{\trace(\Sigma')t}\right] \ge \Pr\left[\norm{X'}^2\ge \trace(\Sigma') - 2\sqrt{\trace(\Sigma')t}\right] \ge 1- e^{-t}\nonumber
	\end{align}
	Finally we observe that $\trace{\Sigma}\ge \trace(\Sigma')\ge \trace(\Sigma)-\norm{\Sigma}$ and this together with equation above completes the proof.
\end{proof}

\subsection{Moments of sub-exponential random variables}

In this subsection, we give several Lemmas regarding the estimation of the moments of (truncated) sub-exponential random variables.

We start with a reduction Lemma that states that one can control the moments of sub-exponential random variable with the moments of the Gaussian random variable and exponential random variables.

\begin{lemma}\label{lem:moment_reduction}
	Suppose $X$ is a sub-exponential random variable with mean 0 and parameters $(\nu,b)$. Let $X_1\sim \mathcal{N}(0,\nu^2)$ and $X_2$ be a exponential random variable with mean $2b$. Let $p\ge 1$ be an integer and $s \ge 0$. Then, 
	\begin{align}
	\Exp\left[X^p\indicator{X\ge s}\right]\leq s^p (e^{-s^2/(2\nu^2)}+e^{-s/(2b)})  +  p \sqrt{2\pi}\nu \Exp\left[X_1^{p-1}\indicator{X_1\ge s}\right] + 2pb\Exp\left[X_2^{p-1}\indicator{X_2\ge s}\right]\mper\nonumber
	\end{align}
\end{lemma}

\begin{proof}
	Let $p(x)$ be the density of $X$ and let $G(x) = \Pr[X\ge x]$. By Lemma~\ref{lem:sub-exp-property} we have $G(x)\le e^{-x^2/(2\nu^2)}+e^{-x/(2b)}$. Then we have 
	\begin{align}
	\Exp\left[X^p\indicator{X\ge s}\right] & = \int_s^{\infty} x^p p(x) dx = - \int_s^{\infty} x^p dG(x)\nonumber\\
	& = - x^p G(x)\big\vert_s^{\infty} + \int_s^{\infty} px^{p-1}G(x)dx\tag{By integration by parts}\\
	& \le s^p (e^{-s^2/(2\nu^2)}+e^{-s/(2b)})  +  \int_s^{\infty} px^{p-1}(e^{-x^2/(2\nu^2)}+e^{-x/(2b)})dx\nonumber\\
	& = s^p (e^{-s^2/(2\nu^2)}+e^{-s/(2b)})  +  p \sqrt{2\pi}\nu \Exp\left[X_1^{p-1}\indicator{X_1\ge s}\right] + 2pb\Exp\left[X_2^{p-1}\indicator{X_2\ge s}\right]\nonumber\mper\end{align}
	
\end{proof}

Next we bound the $p$-th moments of a sub-exponential random variable with zero mean. 

\begin{lemma}\label{lem:sub-exponential-mean-zero-moments}
	Let $p$ be an even integer and let $X$ be a sub-exponential random variable with mean 0 and parameters $(\nu,b)$. Then we have that 
	\begin{align}
	\Exp\left[|X|^p\right]\lesssim \nu^p p!! + (2b)^p p!
	\end{align}
\end{lemma}
\begin{proof}
	Let $X_1\sim \mathcal{N}(0,\nu^2)$ and $X_2$ be a exponential random variable with mean $2b$. By Lemma~\ref{lem:moment_reduction},  we have that 
	\begin{align}
	\Exp\left[X^p \indicator{X\ge 0} \right]& \leq  p \sqrt{2\pi}\nu \Exp\left[X_1^{p-1}\indicator{X_1\ge 0}\right] + 2pb\Exp\left[X_2^{p-1}\right]\mper\nonumber \\
	& \lesssim  \nu^p p!! + (2b)^p p! \mper\nonumber
	\end{align}
	For the negative part we can obtain the similar bound, $	\Exp\left[(-X)^p\indicator{X\le 0}\right]\lesssim \nu^p p!! + (2b)^p p!$. These together complete the proof. 
\end{proof}

The Lemma above implies the bound for the moments of sub-exponential variable with arbitrary mean via Holder inequality.

\begin{lemma}\label{lem:sub-exp-before_truncation-moments}

		Let $X$ be a sub-exponential random variable with mean $\mu$ and parameters $(\nu,b)$. Let $p$ be an integer. 		Then for any $\epsilon\in (0,1)$ we have, 
		\begin{align}
		\Exp |X|^p  \lesssim \frac{1}{\epsilon^{p-1}} |\mu|^p + \frac{1}{(1-\epsilon)^{p-1}} \left(\nu^p p!! + (2b)^p p!\right)\mper \nonumber
		\end{align}
\end{lemma}

\begin{proof}
		Write $X = \mu + Z$ where $Z$ has mean 0. Then we have that for any $\epsilon \in (0,1)$, by Holder inequality, 
		\begin{align}
		\Exp\left[|X|^p \right] = \Exp\left[(\mu +Z)^p\right] & \le \frac{1}{\epsilon^{p-1}} |\mu|^p + \frac{1}{(1-\epsilon)^{p-1}} \Exp\left[|Z|^p\right] \mper\nonumber\\
		& \lesssim\frac{1}{\epsilon^{p-1}} |\mu|^p + \frac{1}{(1-\epsilon)^{p-1}} \left(\nu^p p!! + (2b)^p p!\right)\mper \tag{by Lemma~\ref{lem:sub-exponential-mean-zero-moments}}
		\end{align}
\end{proof}

The next few Lemmas are to bound the moments of a truncated sub-exponential random variable. We use Lemma~\ref{lem:moment_reduction} to reduce the estimates to Gaussian and exponential cases. Thus, we first bound the moment of a truncated Gaussian. 
\begin{claim}\label{claim:truncated_gaussian_moment}
	Suppose $X\sim \N(0,1)$. Then, as $p\rightarrow \infty$, we have for $s \ge \sqrt{p}$, $\Exp\left[|X|^p\indicator{|X|\ge s}\right]\le (1+o_p(1))^p \cdot p \cdot e^{-s^2/2}s^p\nonumber$
\end{claim}

\begin{proof}
	We assume $p$ to be an odd number for simplicity. The even case can be bounded by the odd case straightforwardly. Let $\Gamma(\cdot,\cdot)$ be the upper incomplete Gamma function, we have, 
	\begin{align}
	\int_{s}^{\infty} X^p e^{-x^2/2}dx  & = -2^{\frac{p-1}{2}}\cdot \Gamma(\frac{p+1}{2}, x^2/2)\Big\vert_{s}^{\infty} \tag{since $\frac{\partial \Gamma (s,x)}{\partial x} = -x^{s-1}e^{-x}$~\cite{wiki:incompleteGamma}}\\
	& = 2^{\frac{p-1}{2}}\cdot \Gamma(\frac{p+1}{2}, s^2/2) \nonumber\\
	& = 2^{\frac{p-1}{2}} ((p-1)/2)! \cdot e^{-s^2/2}\sum_{k=0}^{(p-1)/2} \frac{1}{k!}\left(\frac{s^2}{2}\right)^k\tag{see equation~(2) of~\cite{wolfram:incompleteGamma}}\\
	& \le (1+o_p(1))^p \cdot 2^{\frac{p-1}{2}} ((p-1)/2)! \cdot e^{-s^2/2}\cdot (p-1)/2\cdot \left(\frac{s^2e}{p-1} \right)^{(p-1)/2}\tag{since $k! \ge e(k/e)^k$ and $\frac{1}{k!}\left(\frac{s^2}{2}\right)^k$ is maximized at $k = (p-1)/2$ when $s^2\ge p$} \\
	& \le (1+o_p(1))^p \cdot p \cdot e^{-s^2/2}s^p\mper\nonumber
	\end{align}
\end{proof}
The next Claim estimates the moments of a truncated exponential random variable. 
\begin{claim}\label{claim:truncated_exponential_moments}
	Suppose $X$ has exponential distribution with mean 1 (that is, the density is $p(x) = e^{-x}$). Then, as $p\rightarrow \infty$, for any $s\ge (1+\epsilon)p$ and $\epsilon\in (0,1)$, we have $\Exp\left[X^p \indicator{X\ge s}\right]\lesssim e^{-s}s^p/\epsilon$
\end{claim}
\begin{proof}
	We have that 
	\begin{align}
	\int_s^{\infty} e^{-x}x^pdx & = - e^{-x}x^p\sum_{k=0}^{p} \frac{p!}{(p-k)!} x^{-k}\Big\vert_{s}^{\infty}\nonumber\\
	&  = e^{-s}s^p\sum_{k=0}^{p} \frac{p!}{(p-k)!} s^{-k} \nonumber\\
	& \le e^{-s}s^p \sum_k \left(\frac{1}{1+\epsilon}\right)^k\tag{Since $s\ge (1+\epsilon)p$}\\
	& \lesssim e^{-s}s^p/\epsilon\mper\nonumber
	\end{align}
\end{proof}

Combing the two Lemmas and use Lemma~\ref{lem:moment_reduction} we obtain the bounds for sub-exponential random variables. 
\begin{lemma}\label{lem:sub_exponential_moments} \Tnote{This is updated -- check the downstream application}
	Suppose $X$ is a sub-exponential random variable with mean $\mu$ and parameters $(\nu,b)$. Then, for integer $p\ge 2$, real numbers $\eta,\epsilon \in (0,1)$ and $s\ge \max\{\nu \sqrt{p}, 2bp(1+\eta)\}$, we have 
	\begin{align}
		\Exp\left[|X|^p\indicator{|X|\ge \mu + s}\right] \le \frac{1}{\epsilon^{p-1}} |\mu|^p  +  \frac{1}{(1-\epsilon)^{p-1}} s^p \left((1+\zeta_p)^p p^2 e^{-s^2/(2\nu^2)}  +  p e^{-s/(2b)} /\eta\right)\mper \nonumber
	\end{align}
	where $\zeta_p \rightarrow 0$ as $p\rightarrow \infty$. 
\end{lemma}

\begin{proof}
	Write $X = \mu + Z$ where $Z$ has mean 0. Then we have that for any $\epsilon \in (0,1)$, by Holder inequality, 
	\begin{align}
	|X|^p = (\mu +Z)^p \le \frac{1}{\epsilon^{p-1}} |\mu|^p + \frac{1}{(1-\epsilon)^{p-1}} |Z|^p \mper\label{eqn:39}
	\end{align}
	Therefore it remains to bound $\Exp\left[|Z|^p\right]$. Let $p(z)$ be the density of $Z$. Since $Z$ is sub-exponential, we have that $p(z) \le \max\{e^{-z^2/(2\nu^2)}, e^{-z/(2b)}\} \le e^{-z^2/(2\nu^2)}+e^{-z/(2b)}$. Let $Z_1\sim \mathcal{N}(0,\nu^2)$ and $Z_2$ be an exponential random variable with mean $2b$. Then by Lemma~\ref{lem:moment_reduction}, we have 
	\begin{align}
	\Exp\left[|Z|^p\indicator{|Z|\ge s} \right] 
	& \le  s^p (e^{-s^2/(2\nu^2)}+e^{-s/(2b)})  +  p \sqrt{2\pi}\nu \Exp\left[Z_1^{p-1}\indicator{X_1\ge s}\right] + 2pb\Exp\left[X_2^{p-1}\indicator{X_2\ge s}\right] \nonumber\\
	& \le  s^p (e^{-s^2/(2\nu^2)}+e^{-s/(2b)})  +  p \sqrt{2\pi}\nu^p \Exp\left[(Z_1/\nu)^{p-1}\indicator{X_1/\nu\ge s/\nu}\right] \nonumber\\
	& + p(2b)^p\Exp\left[(X_2/2b)^{p-1}\indicator{X_2/(2b)\ge s/(2b)}\right]\nonumber\\
	& \le s^p (e^{-s^2/(2\nu^2)}+e^{-s/(2b)})  + p^2\sqrt{2\pi}e^{-s^2/(2\nu^2)}s^p (1+o_p(1))^p   + pe^{-s/(2b)}s^p /\eta\nonumber\\
	& \lesssim  p^2\sqrt{2\pi}e^{-s^2/(2\nu^2)}s^p (1+o_p(1))^p   + pe^{-s/(2b)}s^p /\eta\mper\nonumber 	\end{align}
	Combing inequality above and equation~\eqref{eqn:39} we complete the proof. 
\end{proof}

Finally, the following is a helper claim that is used in the proof of Claim~\ref{claim:truncated_exponential_moments}. 
\begin{claim}\label{claim:integral}
	For any $\mu \in \R$ and $\beta > 0$, we have that 
	\begin{align}
	\int e^{-\beta x} (\mu+x)^d dx = - \frac{e^{-\beta x}(\mu+x)^d}{\beta}\sum_{k=0}^{d} \frac{d!}{(d-k)!}\left((\mu+x)\beta\right)^{-k} \label{eqn:18}
	\end{align}
\end{claim}

\begin{proof}
	Let $F(x)$ denote the RHS of~\eqref{eqn:18}. We differentiate $F(x)$ and obtain that 
	\begin{align}
	\frac{dF(x)}{dx} & = e^{-\beta x}(\mu+x)^d\sum_{k=0}^{d} \frac{d!}{(d-k)!}\left((\mu+x)\beta\right)^{-k} -\frac{e^{-\beta x}}{\beta}\sum_{k=0}^{d-1} \frac{d!}{(d-k-1)!}\frac{(\mu+x)^{d-k-1}}{\beta^k}\nonumber\\
	& =  e^{-\beta x}\sum_{k=0}^{d} \frac{d!}{(d-k)!}\frac{(\mu+x)^{d-k}}{\beta^k}- e^{-\beta x}\sum_{k=1}^{d} \frac{d!}{(d-k)!}\frac{(\mu+x)^{d-k}}{\beta^k}\nonumber\\
	&= e^{-\beta x} (1+x)^d \mper\nonumber
	\end{align}
	Therefore, the claim follows from integration both hands of the equation above. 
\end{proof}

\subsection{Auxiliary Lemmas}

\begin{lemma}\label{lem:zwnorm}
	Let $\bartau$ be unit vector in $\R^n$, $w\in \R^n$,  and $z \sim \N(0,\sigma^2\cdot (\Id_n-\bar{q}\bar{q}^{\top}))$. Then $\norm{z\odot w}^{2}$ is a sub-exponential random variable with parameter $(\nu,b)$ satisfying $\nu\lesssim \sigma^2\|w\|_4^2$ and $b\lesssim \sigma^2\|w\|_{\infty}^2$. Moreover, it has mean $\sigma^2(\|w\|^2 - \inner{\bar{q}^{\odot 2},w^{\odot 2}})$. 
\end{lemma}

\begin{proof}
	We have that $z\odot w$ is a Gaussian random variable with  covariance matrix $\Sigma = \sigma^2 P_{\bar{q}} \diag(w^{\odot 2})P_{\bar{q}}$. Then, $z\odot w$ can be written as $\Sigma^{1/2}x$ where $x\sim \N(0,\Id_n)$. Then by Lemma~\ref{lem:quadratic_Gaussian}, we have that $\|z\odot w\|^2 = \|\Sigma^{1/2}x\|^2$ is  a sub-exponential random variable with parameter $(\nu,b)$ satisfying \begin{align}
	\nu^2 \lesssim \trace(\Sigma^2) & = \sigma^4\trace(P_{\bar{q}}\diag(w^{\odot 2})P_{\bar{q}}\diag(w^{\odot 2})P_{\bar{q}}) \nonumber\\
	& \le \sigma^4\trace(\diag(w^{\odot 2})P_{\bar{q}}\diag(w^{\odot 2})) \nonumber\\& = \sigma^4\trace(\diag(w^{\odot 4})P_{\bar{q}}) \nonumber\\
& \le \sigma^4\|w\|_4^4\mper\nonumber
	\end{align}
	Here at the second and forth line we both used the fact that $\trace(PA)\le \trace(A)$ holds for any symmetric PSD matrix and projection matrix $P$. 
	
	Moreover, we have
	$b\lesssim \|\Sigma\|\le \sigma^2\max_k w_k^2
	$. Finally, we have the mean of $\|z\odot w\|^2$ is $\trace(\Sigma) = \sigma^2(\|w\|^2 - \inner{\bar{q}^{\odot 2},w^{\odot 2}})$, which completes the proof.  
\end{proof}

\begin{claim}[folklore]\label{claim:factorial}
	We have that $e\left(\frac{n}{e}\right)^n \le n!\le e\left(\frac{n+1}{e}\right)^{n+1}$. It follows that $$ 2e\left(\frac{2(n-1)}{e}\right)^{n-1}\le (2n-2)!!\le (2n-1)!!\le (2n)!! = 2^n n! \le \frac{e}{2}\left(\frac{2(n+1)}{e}\right)^{n+1}\mper$$
\end{claim}

\begin{claim}
	For any $\epsilon \in (0,1)$ and $A,B\in \R$, we have $(A+B)^d \le \max\{(1+\epsilon)^d|A|^d, (1/\epsilon+1)^d |B|^d\}$ 
\end{claim}

\begin{claim} \label{clm:6moment}
For any vectors $a,b$, and any $\epsilon>0$, we have $\|a+b\|_6^6 \le (1+\epsilon) \|a\|_6^6 + O(1/\epsilon^5) \|b\|_6^6$.
\end{claim}

\begin{proof}
By Cauchy-Schwartz we know we only need to prove this when $a,b$ are real numbers instead of vectors. For numbers $a,b$, $(a+b)^6 = a^6 + 6a^5b+ 15a^4b^2 + 20a^3b^3+15a^2b^4+6ab^5+b^6$. All the intermediate terms $a^ib^j (i+j = 6)$ can be bounded by $\epsilon/56 \cdot a^6 + O(1/\epsilon^5) b^6$, therefore we know $(a+b)^6 \le (1+\epsilon) a^6 + O(1/\epsilon^5)b^6$.
\end{proof}

\begin{lemma}\label{lem:1}
	For any vector $z \in \R^n$, and any partition of $[n]$ into two subsets $S$ and $L$, and any $\eta \in (0,1)$, we have that, 
	\begin{align}
	Q(z)^d (\sixmoments{z})^{-d/2}
	& 	\le \frac{1}{\eta^{d/2-1}}\left(\frac{Q(z_S)^2}{\sixmoments{z_S}} \right)^{d/2}+ \frac{1}{(1-\eta)^{d/2-1}}\left( \frac{Q(z_L)^2}{\sixmoments{z_L}}\right)^{d/2}\mper \nonumber
	\end{align}
\end{lemma}

\begin{proof}
	By Cauchy-Schwarz inequality, we have that 
	\begin{align}
	\left(\Norm{z}_4^4-\Norm{z}^2\right)^2 (\Norm{z}_6^6)^{-1} = \frac{(Q(z_S) + Q(z_L))^2}{\sixmoments{z_S}+\sixmoments{z_L}}& \le \frac{Q(z_S)^2}{\sixmoments{z_S}} + \frac{Q(z_L)^2}{\sixmoments{z_L}}\mper \nonumber
	\end{align}
	
	Therefore, we obtain that 
	\begin{align}
	\left(\Norm{z}_4^4-\Norm{z}^2\right)^d (\Norm{z}_6^6)^{-d/2} & \le \left(\frac{Q(z_S)^2}{\sixmoments{z_S}} + \frac{Q(z_L)^2}{\sixmoments{z_L}}\right)^{d/2}\nonumber \\
	& 	\le \frac{1}{\eta^{d/2}}\left(\frac{Q(z_S)^2}{\sixmoments{z_S}} \right)^{d/2}+ \frac{1}{(1-\eta)^{d/2}}\left( \frac{Q(z_L)^2}{\sixmoments{z_L}}\right)^{d/2} \mper\tag{by Holder inequality}
	\end{align}
\end{proof}

%% file: rip_stripped.tex
\subsection{Restricted Isometry Property and Corollaries}
\label{sec:rip}
In this section we will describe the Restricted Isometry Property (RIP), which was introduced in \cite{candes2005decoding}. This is the crucial tool that we used in Section~\ref{sec:local}.

\begin{definition}\label{def:rip}[RIP property] Suppose $A \in \R^{d\times n}$ is a matrix whose columns have unit norm. We say the matrix $A$ satisfies the $(k, \delta)$-RIP property, if for any subset $S$ of columns of size at most $k$, let $A_S$ be the $d\times|S|$ matrix with columns in $S$, we have
$$
1-\delta \le \|A_S\| \le 1+\delta.
$$

In our case the random components $a_i$'s do not have unit norm, we abuse notation and say $A$ is RIP if the column normalized matrix satisfy the RIP property.
\end{definition}

Intuitively RIP condition means the norm of $Ax$ is very close to $\|x\|$, if the number of nonzero entries in $x$ is at most $k$. From \cite{} we know a random matrix is RIP with high probability:

\begin{theorem}[\cite{candes2008restricted}]~\label{thm:RIP} For any constant $\delta > 0$, when $k = d/\Delta\log n$ for large enough constant $\Delta$, with high probability a random Gaussian matrix is $(k,\delta)$-RIP.
\end{theorem}

In our analysis, we mostly use the RIP property to show any given vector $x$ cannot have large correlation with many components $a_i$'s. 

\subsection{Concentration Inequalities}

We first show if we only look at the components with small correlations with $x$, the function value cannot be much larger than $3n$.

\begin{lemma}
\label{lem:degree4}
With high probability over $a_i$'s, for any unit vector $x$, for any threshold $\tau > 1$
$$
\sum_{i=1}^n \indicator{|\inner{a_i,x}| \le \tau} \inner{a_i,x}^4 \le 3n + O((\sqrt{nd}+d\tau^4)\log d).
$$
\end{lemma}

\begin{proof}
Let $X_i = \indicator{|\inner{a_i,x}| \le \tau} \inner{a_i,x}^4 - \E[\indicator{|\inner{a_i,x}| \le \tau} \inner{a_i,x}^4]$. Notice that $\E[\indicator{|\inner{a_i,x}| \le \tau} \inner{a_i,x}^4] \le \E[\inner{a_i,x}^4] = 3$. Clearly $\E[X_i] = 0$ and $X_i \le \tau^4$. We can apply Bernstein's inequality, and we know for any $x$,

$$
\Pr[\sum_{i=1}^n X_i \ge t] \le \exp\left(-\frac{t^2}{15nd+\tau^4t}\right).
$$

When $t \ge C((\sqrt{nd}+d\tau^4)\log d)$ for large enough constant, the probability is smaller than $\exp(-C'd\log d)$ for large constant $C'$, so we can union bound over all vectors in an $\epsilon$-net with $\epsilon = 1/d^3$. This is true even if we set the threshold to $2\tau$. 

For any $x$ not in the $\epsilon$-net, let $x'$ be its closest neighbor in $\epsilon$-net, when $a_i$'s have norm bounded by $O(\sqrt{d})$ (which happens with high probability), it is easy to show whenever $|\inner{x,a_i}|\le \tau$ we always have $|\inner{x',a_i}|\le 2\tau$. Therefore the sum for $x$ cannot be much larger than the sum for $x'$.
\end{proof}

Using very similar techniques we can also show the contribution to the gradient for these small components is small

\begin{lemma}
\label{lem:degree3}
With high probability over $a_i$'s, for any unit vector $x$, for any threshold $\tau > 1$
$$
\|\sum_{i=1}^n \indicator{|\inner{a_i,x}| \le \tau} \inner{a_i,x}^3 a_i\| \le O(n + (\sqrt{nd}+d\tau^4)\log d).
$$
\end{lemma}

\begin{proof}
Let $v = \sum_{i=1}^n \indicator{|\inner{a_i,x}| \le \tau} \inner{a_i,x}^3 a_i$. 
First notice that the inner-product $\inner{v,x}$ is exactly the sum we bounded in Lemma~\ref{lem:degree4}. Therefore we only need to bound the norm in the orthogonal direction of $x$. Since $a_i$'s are Gaussian we know $\inner{a_i,x}$ and $P_x a_i$ are independent. So we can apply vector Bernstein's inequality, and we know

$$
\Pr[\|\sum_{i=1}^n \indicator{|\inner{a_i,x}| \le \tau} \inner{a_i,x}^3 P_x a_i \| \ge t] \le d\exp\left(-\frac{t^2}{15nd+\tau^4t}\right).
$$

By the same $\epsilon$-net argument as Lemma~\ref{lem:degree4} we can get the desired bound.
\end{proof}

Finally we prove similar conditions for the tensor $T$ applied to the vector $x$ twice.

\begin{lemma}
\label{lem:degree2}
With high probability over $a_i$'s, for any unit vector $x$, for any threshold $\tau > 1$
$$
\|\sum_{i=1}^n \indicator{|\inner{a_i,x}| \le \tau} \inner{a_i,x}^2 a_ia_i^\top\| \le O(n + (\sqrt{nd}+d\tau^4)\log d).
$$
\end{lemma}

\begin{proof}
Let $H = \sum_{i=1}^n \indicator{|\inner{a_i,x}| \le \tau} \inner{a_i,x}^2 a_i a_i^\top$. 
First notice that the quadratic form $x^\top H x$ is exactly the sum we bounded in Lemma~\ref{lem:degree4}.
We will bound the spectral norm in the orthogonal direction of $x$, and use the fact that $H$ is a PSD matrix, so $\|H\| \le 2(x^\top H x + \|P_x H P_x|)$ (this is basically the fact that $(a+b)^2 \le 2a^2+2b^2$).  Since $a_i$'s are Gaussian we know $\inner{a_i,x}$ and $P_x a_i$ are independent. So we can apply matrix Bernstein's inequality, because $\E[P_x a_i a_i^\top P_x] = P_x$, we know

$$
\Pr[\|\sum_{i=1}^n \indicator{|\inner{a_i,x}| \le \tau} \inner{a_i,x}^2 (P_x a_i a_i^\top P_x - P_x) \| \ge t] \le d\exp\left(-\frac{t^2}{15nd+\tau^4t}\right).
$$

By the same $\epsilon$-net argument as Lemma~\ref{lem:degree4} we can get the desired bound.
\end{proof}

%% file: proofproofoverview_stripped.tex
\section{Missing proofs in Section~\ref{sec:proof_sketches}}\label{sec:proof:proof_sketches}

\begin{proof}[Proof of Lemma~\ref{lem:G_0}]
	Let $A = [a_1,\dots, a_n]$. First of all by standard random matrix theory (see, e.g., ~[Section 3.3]\cite{ledoux2013probability} or ~\cite[Theorem 2.6]{rudelson2010non}), we have that with high probability, $\Norm{A}\le \sqrt{n}+1.1\sqrt{d}$. Therefore, we have that with high probability $\sum\inner{a_i,x}^2 \le \Norm{A}\le n + 3\sqrt{nd}$. By the RIP property of random matrix $A$  (see~\cite[Theorem 5.2]{baraniuk2008simple} or Theorem~\ref{thm:RIP}), we that with high probability, equation~\eqref{eqn:32} holds. 
	
	Finally, using a straightforward $\epsilon$-net argument and union bound, we have that for every unit vector $x$, $\sum_{i=1}^{n}\inner{a_i,x}^4\ge 15n- O(\sqrt{nd}\log d)$. Therefore, when $1/\delta^2 \cdot d\log^2 d \le n$, equation~\eqref{eqn:29} also holds. 
\end{proof}

%% file: main_stripped.bbl
\newcommand{\etalchar}[1]{$^{#1}$}
\begin{thebibliography}{AGMM15}

\bibitem[AA{\etalchar{+}}13]{auffinger2013complexity}
Antonio Auffinger, Gerard~Ben Arous, et~al.
\newblock Complexity of random smooth functions on the high-dimensional sphere.
\newblock {\em The Annals of Probability}, 41(6):4214--4247, 2013.

\bibitem[AA{\v{C}}13]{auffinger2013random}
Antonio Auffinger, G{\'e}rard~Ben Arous, and Ji{\v{r}}{\'\i} {\v{C}}ern{\`y}.
\newblock Random matrices and complexity of spin glasses.
\newblock {\em Communications on Pure and Applied Mathematics}, 66(2):165--201,
  2013.

\bibitem[AFH{\etalchar{+}}12]{SpectralLDA}
Anima Anandkumar, Dean~P. Foster, Daniel Hsu, Sham~M. Kakade, and Yi-Kai Liu.
\newblock A spectral algorithm for latent {D}irichlet allocation.
\newblock In {\em Advances in Neural Information Processing Systems 25}, 2012.

\bibitem[AGJ15]{anandkumar2015learning}
Animashree Anandkumar, Rong Ge, and Majid Janzamin.
\newblock Learning overcomplete latent variable models through tensor methods.
\newblock In {\em Proceedings of the Conference on Learning Theory (COLT),
  Paris, France}, 2015.

\bibitem[AGJ16]{anandkumar2016analyzing}
Anima Anandkumar, Rong Ge, and Majid Janzamin.
\newblock Analyzing tensor power method dynamics in overcomplete regime.
\newblock {\em JMLR}, 2016.

\bibitem[AGMM15]{rgDict2}
Sanjeev Arora, Rong Ge, Tengyu Ma, and Ankur Moitra.
\newblock Simple, efficient and neural algorithms for sparse coding.
\newblock In {\em Proceedings of The 28th Conference on Learning Theory}, 2015.

\bibitem[AHK12]{AHK12}
Anima Anandkumar, Daniel Hsu, and Sham~M. Kakade.
\newblock A method of moments for mixture models and hidden {M}arkov models.
\newblock In {\em COLT}, 2012.

\bibitem[AMS07]{absil2007optimization}
P.A. Absil, R.~Mahony, and R.~Sepulchre.
\newblock {\em Optimization Algorithms on Matrix Manifolds}.
\newblock Princeton University Press, 2007.

\bibitem[ASS15]{2015arXiv150505729A}
H.~{Abo}, A.~{Seigal}, and B.~{Sturmfels}.
\newblock {Eigenconfigurations of Tensors}.
\newblock {\em ArXiv e-prints}, May 2015.

\bibitem[AT09]{adler2009random}
Robert~J Adler and Jonathan~E Taylor.
\newblock {\em Random fields and geometry}.
\newblock Springer Science \& Business Media, 2009.

\bibitem[BAC16]{2016arXiv160508101B}
N.~{Boumal}, P.-A. {Absil}, and C.~{Cartis}.
\newblock {Global rates of convergence for nonconvex optimization on
  manifolds}.
\newblock {\em ArXiv e-prints}, May 2016.

\bibitem[BBV16]{bandeira2016low}
Afonso~S Bandeira, Nicolas Boumal, and Vladislav Voroninski.
\newblock On the low-rank approach for semidefinite programs arising in
  synchronization and community detection.
\newblock {\em arXiv preprint arXiv:1602.04426}, 2016.

\bibitem[BCMV14]{bhaskara2014smoothed}
Aditya Bhaskara, Moses Charikar, Ankur Moitra, and Aravindan Vijayaraghavan.
\newblock Smoothed analysis of tensor decompositions.
\newblock In {\em Proceedings of the 46th Annual ACM Symposium on Theory of
  Computing}, pages 594--603. ACM, 2014.

\bibitem[BDDW08]{baraniuk2008simple}
Richard Baraniuk, Mark Davenport, Ronald DeVore, and Michael Wakin.
\newblock A simple proof of the restricted isometry property for random
  matrices.
\newblock {\em Constructive Approximation}, 28(3):253--263, 2008.

\bibitem[BKS15]{DBLP:conf/stoc/BarakKS15}
Boaz Barak, Jonathan~A. Kelner, and David Steurer.
\newblock Dictionary learning and tensor decomposition via the sum-of-squares
  method.
\newblock In {\em Proceedings of the Forty-Seventh Annual {ACM} on Symposium on
  Theory of Computing, {STOC} 2015, Portland, OR, USA, June 14-17, 2015}, pages
  143--151, 2015.

\bibitem[BNS16]{bhojanapalli2016global}
Srinadh Bhojanapalli, Behnam Neyshabur, and Nathan Srebro.
\newblock Global optimality of local search for low rank matrix recovery.
\newblock {\em arXiv preprint arXiv:1605.07221}, 2016.

\bibitem[Can08]{candes2008restricted}
Emmanuel~J Candes.
\newblock The restricted isometry property and its implications for compressed
  sensing.
\newblock {\em Comptes Rendus Mathematique}, 346(9):589--592, 2008.

\bibitem[Cha96]{Chang96}
Joseph~T. Chang.
\newblock Full reconstruction of {M}arkov models on evolutionary trees:
  Identifiability and consistency.
\newblock {\em Mathematical Biosciences}, 137:51--73, 1996.

\bibitem[CHM{\etalchar{+}}15]{choromanska2015loss}
Anna Choromanska, Mikael Henaff, Michael Mathieu, G{\'e}rard~Ben Arous, and
  Yann LeCun.
\newblock The loss surfaces of multilayer networks.
\newblock In {\em AISTATS}, 2015.

\bibitem[CLA09]{comon2009tensor}
P.~Comon, X.~Luciani, and A.~De Almeida.
\newblock Tensor decompositions, alternating least squares and other tales.
\newblock {\em Journal of Chemometrics}, 23(7-8):393--405, 2009.

\bibitem[CS13]{cartwright2013number}
Dustin Cartwright and Bernd Sturmfels.
\newblock The number of eigenvalues of a tensor.
\newblock {\em Linear algebra and its applications}, 438(2):942--952, 2013.

\bibitem[CT05]{candes2005decoding}
Emmanuel~J Candes and Terence Tao.
\newblock Decoding by linear programming.
\newblock {\em IEEE transactions on information theory}, 51(12):4203--4215,
  2005.

\bibitem[DLCC07]{de2007fourth}
L.~De~Lathauwer, J.~Castaing, and J.-F. Cardoso.
\newblock Fourth-order cumulant-based blind identification of underdetermined
  mixtures.
\newblock {\em Signal Processing, IEEE Transactions on}, 55(6):2965--2973,
  2007.

\bibitem[DPG{\etalchar{+}}14]{dauphin2014identifying}
Yann~N Dauphin, Razvan Pascanu, Caglar Gulcehre, Kyunghyun Cho, Surya Ganguli,
  and Yoshua Bengio.
\newblock Identifying and attacking the saddle point problem in
  high-dimensional non-convex optimization.
\newblock In {\em Advances in neural information processing systems}, pages
  2933--2941, 2014.

\bibitem[GHJY15]{ge2015escaping}
Rong Ge, Furong Huang, Chi Jin, and Yang Yuan.
\newblock Escaping from saddle points---online stochastic gradient for tensor
  decomposition.
\newblock In {\em Proceedings of The 28th Conference on Learning Theory}, pages
  797--842, 2015.

\bibitem[GLM16]{ge2016matrix}
Rong Ge, Jason~D Lee, and Tengyu Ma.
\newblock Matrix completion has no spurious local minimum.
\newblock {\em arXiv preprint arXiv:1605.07272}, 2016.

\bibitem[GM15]{ge2015decomposing}
Rong Ge and Tengyu Ma.
\newblock Decomposing overcomplete 3rd order tensors using sum-of-squares
  algorithms.
\newblock {\em arXiv preprint arXiv:1504.05287}, 2015.

\bibitem[GVX13]{fourierpca}
N.~Goyal, S.~Vempala, and Y.~Xiao.
\newblock Fourier pca.
\newblock {\em arXiv preprint arXiv:1306.5825}, 2013.

\bibitem[H{\aa}s90]{haastad1990tensor}
Johan H{\aa}stad.
\newblock Tensor rank is np-complete.
\newblock {\em Journal of Algorithms}, 11(4):644--654, 1990.

\bibitem[HK13]{HK13-mog}
Daniel Hsu and Sham~M. Kakade.
\newblock Learning mixtures of spherical {G}aussians: moment methods and
  spectral decompositions.
\newblock In {\em Fourth Innovations in Theoretical Computer Science}, 2013.

\bibitem[HKZ12a]{HKZ12}
Daniel Hsu, Sham~M. Kakade, and Tong Zhang.
\newblock A spectral algorithm for learning hidden {M}arkov models.
\newblock {\em Journal of Computer and System Sciences}, 78(5):1460--1480,
  2012.

\bibitem[HKZ12b]{hsu2012tail}
Daniel Hsu, Sham~M Kakade, and Tong Zhang.
\newblock A tail inequality for quadratic forms of subgaussian random vectors.
\newblock {\em Electron. Commun. Probab}, 17(52):1--6, 2012.

\bibitem[HL13]{hillar2013most}
Christopher~J Hillar and Lek-Heng Lim.
\newblock Most tensor problems are np-hard.
\newblock {\em Journal of the ACM (JACM)}, 60(6):45, 2013.

\bibitem[HM16]{DBLP:journals/corr/HardtM16}
Moritz Hardt and Tengyu Ma.
\newblock Identity matters in deep learning.
\newblock {\em CoRR}, abs/1611.04231, 2016.

\bibitem[HMR16]{DBLP:journals/corr/HardtMR16}
Moritz Hardt, Tengyu Ma, and Benjamin Recht.
\newblock Gradient descent learns linear dynamical systems.
\newblock {\em CoRR}, abs/1609.05191, 2016.

\bibitem[HSSS16]{DBLP:conf/stoc/HopkinsSSS16}
Samuel~B. Hopkins, Tselil Schramm, Jonathan Shi, and David Steurer.
\newblock Fast spectral algorithms from sum-of-squares proofs: tensor
  decomposition and planted sparse vectors.
\newblock In {\em Proceedings of the 48th Annual {ACM} {SIGACT} Symposium on
  Theory of Computing, {STOC} 2016, Cambridge, MA, USA, June 18-21, 2016},
  pages 178--191, 2016.

\bibitem[{Kaw}16]{Kawaguchi}
K.~{Kawaguchi}.
\newblock {Deep Learning without Poor Local Minima}.
\newblock {\em ArXiv e-prints}, May 2016.

\bibitem[KM11]{kolda2011shifted}
Tamara~G Kolda and Jackson~R Mayo.
\newblock Shifted power method for computing tensor eigenpairs.
\newblock {\em SIAM Journal on Matrix Analysis and Applications},
  32(4):1095--1124, 2011.

\bibitem[LM00]{laurent2000}
B.~Laurent and P.~Massart.
\newblock Adaptive estimation of a quadratic functional by model selection.
\newblock {\em Ann. Statist.}, 28(5):1302--1338, 10 2000.

\bibitem[LSJR16]{DBLP:conf/colt/LeeSJR16}
Jason~D. Lee, Max Simchowitz, Michael~I. Jordan, and Benjamin Recht.
\newblock Gradient descent only converges to minimizers.
\newblock In {\em Proceedings of the 29th Conference on Learning Theory, {COLT}
  2016, New York, USA, June 23-26, 2016}, pages 1246--1257, 2016.

\bibitem[LT13]{ledoux2013probability}
Michel Ledoux and Michel Talagrand.
\newblock {\em Probability in Banach Spaces: isoperimetry and processes}.
\newblock Springer Science \& Business Media, 2013.

\bibitem[MR06]{MR06}
Elchanan Mossel and S\'{e}bastian Roch.
\newblock Learning nonsingular phylogenies and hidden {M}arkov models.
\newblock {\em Annals of Applied Probability}, 16(2):583--614, 2006.

\bibitem[MSS16]{MSS16}
Tengyu Ma, Jonathan Shi, and David Steurer.
\newblock Polynomial-time tensor decompositions with sum-of-squares.
\newblock In {\em FOCS 2016, to appear}, 2016.

\bibitem[NP06]{nesterov2006cubic}
Yurii Nesterov and Boris~T Polyak.
\newblock Cubic regularization of newton method and its global performance.
\newblock {\em Mathematical Programming}, 108(1):177--205, 2006.

\bibitem[NPOV15]{2015arXiv150906569N}
A.~{Novikov}, D.~{Podoprikhin}, A.~{Osokin}, and D.~{Vetrov}.
\newblock {Tensorizing Neural Networks}.
\newblock {\em ArXiv e-prints}, September 2015.

\bibitem[RV10]{rudelson2010non}
Mark Rudelson and Roman Vershynin.
\newblock Non-asymptotic theory of random matrices: extreme singular values.
\newblock {\em arXiv preprint arXiv:1003.2990}, 2010.

\bibitem[SQW15]{sun2015nonconvex}
Ju~Sun, Qing Qu, and John Wright.
\newblock When are nonconvex problems not scary?
\newblock {\em arXiv preprint arXiv:1510.06096}, 2015.

\bibitem[Wai15]{wainwrightBasic}
Martin Wainwright.
\newblock Basic tail and concentration bounds, 2015.

\bibitem[Wei16]{wolfram:incompleteGamma}
Eric~W. Weisstein.
\newblock Incomplete gamma function --- from mathworld--a wolfram web resource,
  2016.

\bibitem[Wik16a]{wiki:incompleteGamma}
Wikipedia.
\newblock Incomplete gamma function --- wikipedia{,} the free encyclopedia,
  2016.
\newblock [Online; accessed 13-September-2016].

\bibitem[Wik16b]{wiki:fixedpoint}
Wikipedia.
\newblock Schauder fixed point theorem --- wikipedia{,} the free encyclopedia,
  2016.
\newblock [Online; accessed 26-May-2016].

\end{thebibliography}
